\pdfoutput=1
\documentclass[10pt,twocolumn,letterpaper]{article}
\usepackage[T1]{fontenc}
\usepackage[utf8]{inputenc}
\usepackage{lmodern}
\usepackage[top=2.5cm,bottom=2.5cm,left=2cm,right=2cm,columnsep=1.2em]{geometry}
\usepackage{amsmath,amssymb,mathtools}
\usepackage{microtype}
\usepackage{graphicx,booktabs,array,tabularx}
\usepackage[table]{xcolor}
\usepackage{caption}
\usepackage{fvextra}
\usepackage{enumitem}
\usepackage{titlesec}
\usepackage{xurl}
\usepackage{stfloats}
\definecolor{linkblue}{HTML}{2E6DAD}
\definecolor{groupgray}{gray}{0.94}
\usepackage[colorlinks=true,linkcolor=linkblue,citecolor=linkblue,urlcolor=linkblue,unicode,hypertexnames=false]{hyperref}
\hypersetup{pdftitle={Does the Mixer Matter? Paired-Seed Ablations of Tiny Bilingual Language Models},pdfauthor={Ayoub Kirouane}}
\DeclareUnicodeCharacter{2212}{\ensuremath{-}}
\DeclareUnicodeCharacter{00B1}{\ensuremath{\pm}}
\DeclareUnicodeCharacter{00D7}{\ensuremath{\times}}
\DeclareUnicodeCharacter{2190}{\ensuremath{\leftarrow}}
\DeclareCaptionType{diagram}[Diagram][List of Diagrams]
\titleformat*{\section}{\large\bfseries}
\titleformat*{\subsection}{\normalsize\bfseries}
\titleformat*{\subsubsection}{\normalsize\bfseries\itshape}
\titlespacing*{\section}{0pt}{1.6ex plus .4ex minus .2ex}{0.8ex}
\titlespacing*{\subsection}{0pt}{1.2ex plus .3ex minus .2ex}{0.5ex}
\titlespacing*{\subsubsection}{0pt}{1ex plus .3ex minus .2ex}{0.4ex}
\setlist{nosep}

\providecommand{\tightlist}{\setlength{\itemsep}{0pt}\setlength{\parskip}{0pt}}

\begin{document}
\twocolumn[{%
\begin{center}
$\vcenter{\hbox{\includegraphics[height=0.9cm]{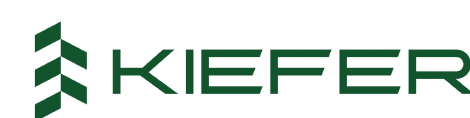}}}$\hspace{1.3cm}$\vcenter{\hbox{\includegraphics[height=1.2cm]{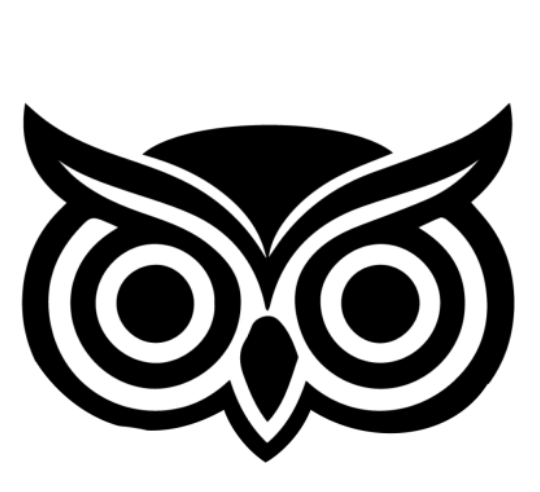}}}$\\[0.9em]
{\fontsize{16}{19}\selectfont\bfseries Does the Mixer Matter? Paired-Seed Ablations of Tiny Bilingual Language Models\par}\vspace{0.7em}
{\fontsize{11}{13}\selectfont Ayoub Kirouane\par}\vspace{0.3em}
{\small Sophea AI Lab, KIEFER SA, Athens, Greece\par}\vspace{0.2em}
{\small\href{mailto:a.kirouane@kiefer.gr}{a.kirouane@kiefer.gr}\par}\vspace{0.3em}
{\footnotesize\color{black!67}November 2026\par}
\end{center}
\vspace{0.4em}
\begin{center}\begin{minipage}{0.86\textwidth}
\centerline{\bfseries Abstract}\vspace{0.4em}
\small We report paired-seed architecture ablations from the development of Sophea Artemis, our in-house Greek--English model family. Every arm trains on the same 3B targets with a shared tokenizer and recipe; models range from 54M to 759M total parameters and roughly 54M to 124M active parameters. The main finding is that, at this tiny scale, stored capacity matters far more than mixer design. On a fixed held-out test, replacing the dense feed-forward blocks of a recurrent/attention hybrid with sixteen routed experts, at nearly the same approximate active parameters, lowers loss by 0.079 (Greek) and 0.112 (English) nats per target, and doubling to 32 experts lowers it by a further 0.022 and 0.031, at every paired seed. Reusing a four-layer block three times recovers about 40--44\% of the gap between a shallow and an independently deep model, consistent with looped-Transformer results. In contrast, the choice among grouped-query attention (GQA), Gated DeltaNet (GDN) hybrids and a Kimi Delta Attention (KDA) hybrid moves loss by only about 0.01--0.02 nats; a 1:1 GDN/GQA hybrid is the one dense variant that improves on GQA in both languages at every held-out seed. In bits per character, Greek and English held-out losses are similar despite a 74.49\%/25.51\% training split, and synthetic mathematics, which comes from a collection outside the training sources, is the hardest Greek test component and gains least from added experts. Cached decoding needs its own qualification: trained models passed cached-versus-full logit checks only under a fixed-reduction inference policy, which itself changes 1.4--2.8\% of next-token argmax choices relative to native evaluation. All results are conditional on one recipe and scale.\par
\end{minipage}\end{center}
\vspace{1.2em}
}]
\section{Introduction}\label{introduction}

Tiny language models can spend their resources in different ways: additional experts increase stored capacity while activating only a subset for each token, reusing a layer block reduces unique weights but repeats computation, and recurrent mixing changes the computation and state needed to process a sequence. Here, tiny means a 768-wide decoder with at most 124M active parameters per token, trained on 3B targets at 2,048-token context; total stored parameters range from 54M to 759M (Table 1).

We compare these choices in the pretraining campaign for Sophea Artemis, an in-house family of Greek--English models. All main comparisons use the same tokenizer, language mixture (74.49\% Greek and 25.51\% English targets), context length and training-target schedule, with three paired seeds per primary comparison. The candidate components are established: GQA shares key/value heads \cite{ref1}; GDN and KDA provide recurrent mixing \cite{ref2,ref3}; sparse experts provide conditional parameter activation \cite{ref4}; and shared-depth Transformers reuse weights across computational steps \cite{ref7,ref8}. The question is which of them matters under one controlled recipe, and the answer is mostly capacity (Figure 2). Our contributions are:

\begin{enumerate}
\def\labelenumi{\arabic{enumi}.}
\tightlist
\item
  \textbf{Capacity versus mixing.} Paired-seed evidence that stored capacity dominates mixer design at this scale. Replacing dense feed-forward blocks with sixteen routed experts at matched approximate active parameters lowers loss by about 0.08--0.11 nats, moving from 16 to 32 experts by a further 0.02--0.03, and shared depth recovers about 40--44\% of the shallow-to-deep gap, while mixer choices move loss by only about 0.01--0.02 nats. The expert-count result confirms the direction expected when routed parameters are scaled at fixed activated computation \cite{ref23,ref24}, and the shared-depth result replicates the looped-Transformer pattern of \cite{ref8}. The mixer result is the least predictable from prior reports: the attention-richer 1:1 hybrid improves on GQA in both languages at every held-out seed, whereas the 3:1 ratio, which Kimi Linear found best at an equal-FLOPs budget \cite{ref3}, shows no consistent difference from GQA.
\item
  \textbf{Greek--English measurement.} Tokenizer fertility and held-out bits per character (BPC), which put Greek and English on one scale, and a breakdown of the Greek test pool by source component.
\item
  \textbf{Cached-inference consistency.} A model can pass synthetic checks yet fail cached-versus-full checks on real text, and a policy that makes the two agree can still change a few percent of next-token choices relative to native evaluation.
\end{enumerate}

\section{Related work}\label{related-work}

\textbf{Attention and recurrent mixing.} We use GQA \cite{ref1}, trained from scratch, as the comparison anchor. GDN introduces gated delta-rule recurrence and evaluates recurrent/attention hybrids \cite{ref2}. Kimi Linear introduces KDA and combines it with multi-head latent attention (MLA) \cite{ref3}; DeepSeek-V2 also provides an established MLA and sparse-expert design \cite{ref5}. Kimi Linear found three KDA layers per MLA layer best at a common FLOPs budget, with 1:1 close behind, and its KDA hybrid outperformed a GDN hybrid \cite{ref3}. Our KDA variant replaces the recurrent blocks of our GDN/GQA expert hybrid and keeps the GDN SiLU output gate instead of the source method's low-rank sigmoid gate, so it is not a reproduction of Kimi Linear; MLA is untested. Our attention layers use the per-head sigmoid output gate of Qiu et al.~\cite{ref20}, and the 3:1 expert hybrid shares its Gated DeltaNet/gated-attention layout, single shared expert and zero-centered normalization with the published Qwen3-Next design \cite{ref21}. Rotary position encoding (RoPE) follows \cite{ref6}. These backbone choices are adopted, not proposed here.

\textbf{Sparse capacity.} Learned sparse expert selection is established by Shazeer et al.~\cite{ref4}. Our normalized sigmoid routing follows DeepSeek-V3 \cite{ref16}, the selection-only bias balancing is the auxiliary-loss-free strategy of Wang et al.~\cite{ref22}, and the always-active shared expert follows DeepSeekMoE \cite{ref23}. Routed scaling studies report gains from adding experts at fixed activated computation \cite{ref24}.

\textbf{Shared computation.} Universal Transformers use recurrent self-attentive computation \cite{ref7}. Saunshi et al.~report that looped Transformers improve perplexity over iso-parameter baselines but recover only roughly 34--50\% of the gap to the iso-FLOP non-looped baseline \cite{ref8}. Takase and Kiyono study earlier strategies for sharing parameters across layers \cite{ref25}, and Monea et al.~study shared memory across recursions \cite{ref9}; we instead keep separate key/value states for each logical layer visit. Lee et al.~\cite{ref12} and LoopMoE \cite{ref13} study looping with sparse experts, which our fixed-pass dense controls do not test.

\textbf{Controlled comparisons and the language setting.} Narang et al.~show that Transformer modifications often fail to transfer across implementations \cite{ref10}, so we restrict claims to one recipe. Borobia et al.~instead ablate components of already trained sub-1B hybrids \cite{ref11}. For Greek, LexisLM reports Greek-only pretraining of compact decoders \cite{ref14}, and the Typakos model card describes a small Greek--English model trained from scratch at 2k context \cite{ref15}. Different tokenizers and evaluations prevent direct loss comparisons with them.

\section{Setup}\label{setup}

\subsection{Models}\label{models}

The main envelope has 12 decoder layers, width 768 and head dimension 128. Dense feed-forward blocks use swish-gated linear units (SwiGLU) with intermediate width 2,304. The standard hybrid repeats three GDN layers followed by one GQA layer; the alternating hybrid uses one GDN layer per attention layer. Mixture-of-experts (MoE) variants use the 3:1 hybrid backbone with top-2 routing, one shared expert and per-expert intermediate width 768. KDA replaces the nine recurrent layers of the 3:1 expert hybrid; it changes the gating family, including channelwise decay and low-rank forget projections, so it is not a scalar-gate-only intervention. The looped model registers one four-layer GQA block and reuses all of its parameters for three passes; the shallow control executes that block once and the deep control uses 12 independently parameterized layers. Table 1 gives exact parameter counts, Figure 1 the controls, and Appendix A the full specification (Equations 1--15, Diagrams 1--6 and Algorithms 1--2). Figures and tables use shorter labels: the GDN/GQA expert hybrids appear as Hybrid MoE16/MoE32 or GDN MoE16/MoE32, the 12-layer recurrent model as GDN12 or all-GDN, and R in Figure 1 marks a scalar-gated GDN layer.

\begin{table*}[t]
\caption{\textbf{Architecture configurations.} {\footnotesize Total (unique) and approximate active parameter counts, with completed training seeds.}}\label{tab:architecture}
\small\begin{tabularx}{\textwidth}{@{}>{\hsize=1.5182\hsize\raggedright\arraybackslash}X>{\hsize=0.8467\hsize\raggedleft\arraybackslash}X>{\hsize=0.8467\hsize\raggedleft\arraybackslash}X>{\hsize=0.7883\hsize\centering\arraybackslash}X@{}}
\toprule
 & \multicolumn{2}{c}{\textbf{Parameters}} & \\
\cmidrule(lr){2-3}
\textbf{Architecture} & \textbf{Total (unique)} & \textbf{Active (approx.)} & \centering\arraybackslash\textbf{Completed seeds}\\
\midrule
\rowcolor{groupgray}\multicolumn{4}{@{\hspace{4pt}}l}{\textbf{Dense backbones}}\\
GQA, 12 independent layers & 114,251,520 & 114,251,520 & 42, 43, 44 (a)\\
GDN, 12 layers & 123,908,496 & 123,908,496 & 42\\
Dense GDN/GQA, 3:1 & 121,494,252 & 121,494,252 & 42, 43, 44\\
Dense GDN/GQA, 1:1 & 119,080,008 & 119,080,008 & 42, 43, 44\\
\rowcolor{groupgray}\multicolumn{4}{@{\hspace{4pt}}l}{\textbf{Sparse expert hybrids}}\\
GDN/GQA, 16 routed experts & 418,913,004 & 121,641,708 & 42, 43, 44\\
GDN/GQA, 32 routed experts & 758,799,084 & 121,789,164 & 42, 43, 44\\
KDA/GQA, 16 routed experts & 420,647,862 & 123,376,566 & 42, 43, 44 (c)\\
\rowcolor{groupgray}\multicolumn{4}{@{\hspace{4pt}}l}{\textbf{Shared-depth controls}}\\
GQA, 4 independent layers & 54,468,352 & 54,468,352 & 42, 43, 44\\
GQA, 4 layers, 3 shared passes & 54,468,352 & 54,468,352 (b) & 42, 43, 44\\
\bottomrule
\end{tabularx}
\par\vspace{4pt}\footnotesize
\begin{tabularx}{\textwidth}{@{}l@{\hspace{5pt}}X@{}}
\textbf{Notes.} & Counts are shown without rounding. Active parameters are approximate and do not measure FLOPs or latency.\\
\textbf{(a)} & Each seed completed in both numerical modes.\\
\textbf{(b)} & Unique active weights. The shared block executes three times (12 logical layer visits); the 12-layer GQA is the independent-depth control.\\
\textbf{(c)} & All three KDA seeds completed and passed endpoint checks.\\
\end{tabularx}
\end{table*}

\begin{figure*}[t]
\centering
\includegraphics[width=1.0\textwidth]{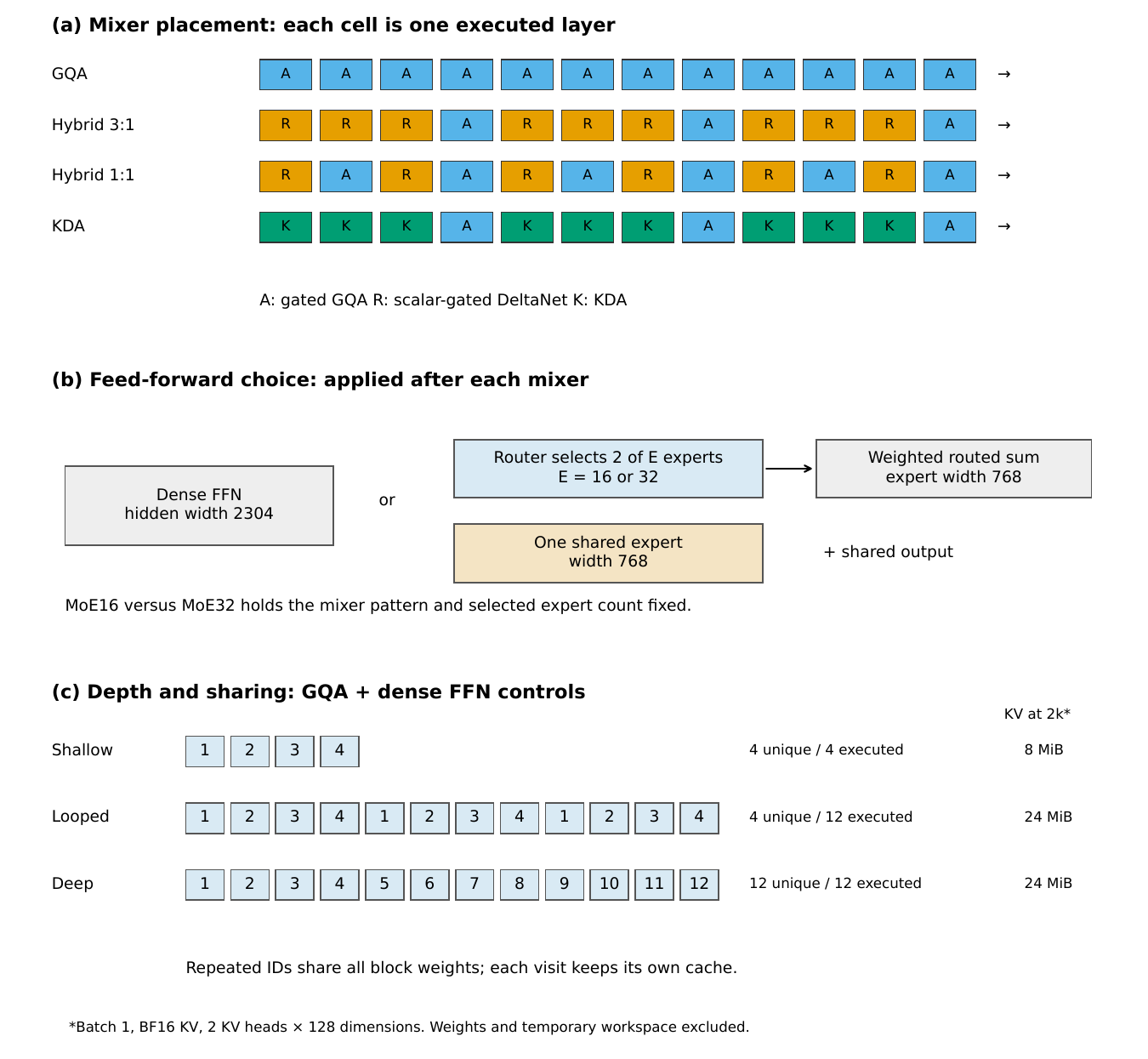}
\caption{Selected architecture controls. Panel (a) shows the configured order of twelve mixer layers. Panel (b) distinguishes dense feed-forward blocks from routed-plus-shared experts; MoE16 and MoE32 use the same 3:1 mixer pattern and top-2 selection. Panel (c) distinguishes shared weights from executed depth and logical KV storage at batch 1, context 2,048 and BF16. Panels describe controlled comparisons, not a factorial sweep of every possible combination. Residual connections, norms and output gates are omitted from this schematic.}\label{fig:1}
\end{figure*}

\subsection{Data}\label{data}

The training corpus combines public-source material with unpublished, locally held Greek--English collections; eight source components contribute through frozen mixture weights. Neither the story collection nor the synthetic-mathematics collection that together supply 51.44\% of Greek held-out targets is among these training sources, so both components are out of distribution for every arm. Every base arm trains on 3B targets with nominal shares of 74.49\% Greek and 25.51\% English. This corresponds to about 24--26 training tokens per parameter for the dense 12-layer models, close to the roughly 20 tokens per parameter of compute-optimal training \cite{ref26}, and about 55 for the four-layer models. The expert models receive about 25 tokens per active parameter but only about 7 (16 experts) and 4 (32 experts) per stored parameter, so their added capacity is heavily undertrained relative to that reference. Largest-remainder quotas allocate complete rows of 2,048 targets, and seeded per-source row selection draws from the full prepared supply with at most one pass per source. Seeds change both initialization and sampled row selection/order, so they are repetitions of the full training procedure, not initialization-only repeats; paired comparisons use matching seeds and quotas. All experiments use the same frozen tokenizer with a vocabulary of 32,000 tokens. Canonical deduplication and exact/normalized 13-word overlap exclusion protect the held-out pools (Appendix B).

\subsection{Training}\label{training}

The native backend uses a fixed TorchTitan version, with PyTorch \texttt{2.16.0.dev20261015+cu132}, bfloat16 (BF16) computation and 32-bit floating-point (FP32) gradient reduction. The shared optimizer policy is AdamW with peak learning rate 0.0003, betas 0.9/0.95, epsilon \(10^{-8}\) and global gradient-norm clipping at 1. Weight decay 0.1 applies to matrix-shaped weights; embeddings and the tied output projection, normalization gains, router weights, recurrent decay parameters and biases are not decayed. Each update contains 8,192 prediction targets (four 2,048-target sequences, no gradient accumulation). Warmup spans 3,663 updates, followed by cosine decay to 10\% of the peak rate (\(3\times10^{-5}\)) at the common endpoint, step 366,211. Linear weights are initialized from \(\mathcal{N}(0,0.02^2)\), except that residual-output projections of dense FFNs and experts use standard deviation \(0.02/\sqrt{2(l+1)}\) for layer index \(l\); normalization gains start at one and biases at zero. Checkpoints occur every 4,096 updates; validation occurs at the first update, every 8,192 updates and the final update. Per-family optimizer tuning has not been performed.

Original family comparisons request deterministic mode off. Shallow and shared-depth experiments request it on, so the 12-layer GQA control was rerun with the same requested mode at all three seeds. We use those matched runs for depth comparisons and the original-mode GQA runs for family comparisons; the reruns are numerical controls, not additional independent seeds.

All 32 selected runs trained on NVIDIA B200 graphics processing units (GPUs) with the same per-run device configuration. Allocated GPU-hours are the device-hours allocated to a run. They include startup, validation and checkpoint work and exclude queue wait. Across the three seeds of each primary group, the sample SD of allocated hours is at most 0.25 hours (MoE16) and below 0.13 hours for every other group, and elapsed validation timestamps match the allocated GPU-hours within 0.05 h for all 32 runs.

\subsection{Evaluation and statistics}\label{evaluation-and-statistics}

\textbf{Validation.} Each language is evaluated on a fixed 131,072-target packed validation subset. We report token-averaged negative log-likelihood (NLL) in nats per target (Equation 15); lower is better. All validation tables use the final scheduled endpoint, not the best validation checkpoint.

\textbf{Held-out test.} A separately registered evaluation scores all 32 selected endpoints on 1,635 Greek documents (1,048,135 targets) and 962 English documents (1,048,000 targets) in document segments of at most 2,048 prediction pairs (Algorithm 2). Greek pools combine synthetic mathematics (33.58\% of targets), stories (17.86\%), web text (45.48\%) and reference and other text (3.08\%); English is educational web text. The synthetic-mathematics and story components come from collections outside the training mixture (Section 3.2). Checkpoint selection, document manifests and scoring were fixed in a written protocol before any test score was computed.

\textbf{Bits per character.} Per-target NLL depends on the tokenizer, so we also report held-out BPC: the summed loss in nats divided by \(\ln 2\) and by the number of predicted characters. Characters are those of the predicted tokens; each document's first token is context, not a target, and the appended EOS contributes loss but no characters. Fertility (characters, UTF-8 bytes and whitespace words per token) is measured on the same held-out documents.

\textbf{Statistics.} For three-seed groups, tables show arithmetic means and sample standard deviations (SD). For every paired contrast we report the number of improving seeds and the mean of candidate-minus-baseline differences, with their sample SD on validation and a document-bootstrap interval on test (Table 14). On the test we also report a paired document bootstrap, added after scoring: 2,000 resamples of test documents, the same resample applied to both arms and all three seeds, with the seed-mean paired difference as the statistic. Its 95\% interval reflects test-document sampling only. The campaign was adaptive: which configurations received seeds 43 and 44 was decided after inspecting seed 42 validation results.

\section{Results}\label{results}

Figure 2 summarizes the study. On the held-out test, the routed-expert models occupy a clearly lower band than every dense model in both languages, shared depth moves the four-layer model about 40\% of the way toward independent depth, and the dense hybrids lie within 0.005 bits per character of GQA (differences of seed means).

\begin{figure*}[t]
\centering
\includegraphics[width=1.0\textwidth]{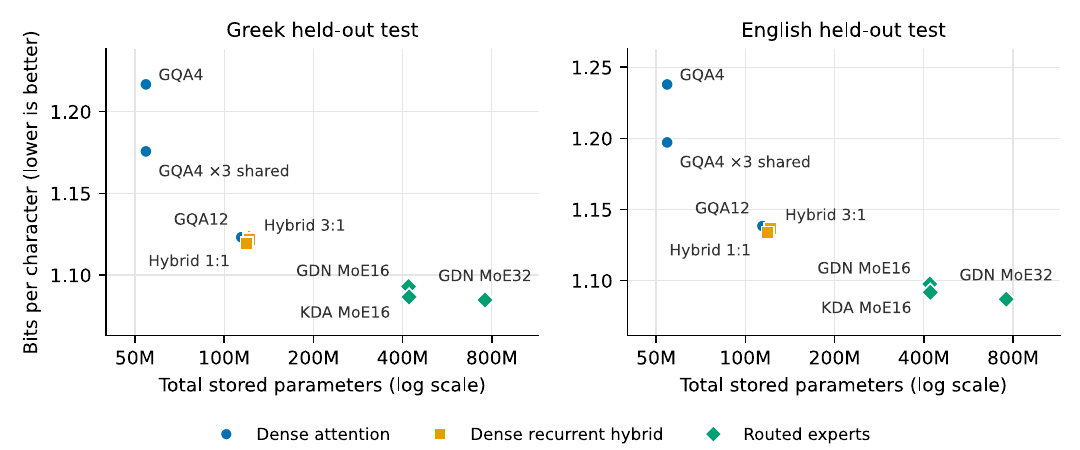}
\caption{Held-out bits per character versus total stored parameters (log scale) for the eight three-seed groups, Greek and English test panels. Colors and markers distinguish dense attention, dense recurrent hybrids and routed-expert models. Points are seed means; vertical bars show the sample SD over three seeds and are mostly smaller than the markers. GQA12 is the original-mode control.}\label{fig:2}
\end{figure*}

\subsection{Routed-expert capacity}\label{routed-expert-capacity}

\begin{table*}[t]
\caption{Routed expert capacity: quality and training cost.}\label{tab:layout4}
\small
\begin{tabularx}{\textwidth}{@{}>{\hsize=1.2500\hsize\raggedright\arraybackslash}X>{\hsize=0.4000\hsize\raggedleft\arraybackslash}X>{\hsize=1.2000\hsize\raggedleft\arraybackslash}X>{\hsize=1.2000\hsize\raggedleft\arraybackslash}X>{\hsize=0.9500\hsize\raggedleft\arraybackslash}X@{}}
\toprule
 & & \multicolumn{2}{c}{\textbf{Validation NLL}} & \\
\cmidrule(lr){3-4}
\textbf{Model} & \centering\arraybackslash \textbf{Seeds} & \centering\arraybackslash \textbf{Greek} & \centering\arraybackslash \textbf{English} & \centering\arraybackslash \textbf{Allocated GPU-hours}\\
\midrule
Hybrid MoE16 & 3 & 3.07222 ± 0.00854 & 3.24020 ± 0.01038 & 16.74\\
Hybrid MoE32 & 3 & 3.04577 ± 0.01138 & 3.20732 ± 0.01011 & 16.86\\
\bottomrule
\end{tabularx}
\par\vspace{3pt}{\footnotesize\raggedright\textbf{Notes.} Three paired seeds per model. NLL is in nats per target (mean ± sample SD; lower is better). GPU-hours are mean allocated training time. Top-2 selection, expert width and the backbone are held fixed.\par}
\end{table*}

The closest control for added routed capacity is the dense 3:1 hybrid, which shares the MoE16 mixer backbone and has nearly the same approximate active parameters (121.49M versus 121.64M). Replacing its dense FFNs with sixteen routed experts plus one shared expert lowers mean validation loss by 0.08232 Greek and 0.11426 English at every paired seed (paired SDs 0.00918 and 0.00405), at 5.14 more allocated GPU-hours. Total stored parameters rise from 121.5M to 418.9M, so this isolates added routed capacity at matched approximate activation, not sparsity at matched total parameters. Against original-mode GQA the gains are similar (−0.08143 Greek and −0.11983 English).

Moving from 16 to 32 routed experts, with the backbone, top-2 selection and expert width fixed, improves both languages at every paired seed (Table 2, Figures 3 and 4). Mean improvements are 0.02645 Greek and 0.03288 English, while observed mean allocation rises by 0.12 hours. Total parameters increase from 418.9M to 758.8M; approximate active parameters change much less. Both expert effects hold on the held-out test (Section 4.4).

\subsection{Shared depth}\label{shared-depth}

\begin{table*}[t]
\caption{Shared versus independent depth.}\label{tab:layout5}
\small
\begin{tabularx}{\textwidth}{@{}>{\hsize=1.2500\hsize\raggedright\arraybackslash}X>{\hsize=0.6500\hsize\raggedleft\arraybackslash}X>{\hsize=1.1500\hsize\raggedleft\arraybackslash}X>{\hsize=1.1500\hsize\raggedleft\arraybackslash}X>{\hsize=0.8000\hsize\raggedleft\arraybackslash}X@{}}
\toprule
 & & \multicolumn{2}{c}{\textbf{Validation NLL}} & \\
\cmidrule(lr){3-4}
\textbf{Model} & \centering\arraybackslash \textbf{Unique params.} & \centering\arraybackslash \textbf{Greek} & \centering\arraybackslash \textbf{English} & \centering\arraybackslash \textbf{Allocated GPU-hours}\\
\midrule
\rowcolor{groupgray}\multicolumn{5}{@{\hspace{4pt}}l}{\textbf{Same unique-weight budget}}\\
Four layers, one pass & 54.47M & 3.41076 ± 0.01104 & 3.65005 ± 0.00749 & 4.29\\
Same four layers, three passes & 54.47M & 3.30205 ± 0.00581 & 3.53414 ± 0.00875 & 8.84\\
\rowcolor{groupgray}\multicolumn{5}{@{\hspace{4pt}}l}{\textbf{Independent-depth control}}\\
Twelve independent layers & 114.25M & 3.15382 ± 0.00427 & 3.36166 ± 0.00560 & 9.28\\
\bottomrule
\end{tabularx}
\par\vspace{3pt}{\footnotesize\raggedright\textbf{Notes.} Three seeds per model, with matched requested numerical mode. Parameters are unique weights in millions; NLL is mean ± sample SD. GPU-hours are mean allocated training time. Shared passes reuse weights but retain separate logical caches.\par}
\end{table*}

All three seeds show the same ordering in both languages (Table 3, Figure 3). Reusing the four-layer block three times improves mean loss over one pass by 0.10871 Greek and 0.11592 English, at an additional 4.55 allocated GPU-hours. It remains worse than independent 12-layer depth by 0.14823 and 0.17248, while saving only 0.44 observed GPU-hours on average. Looping therefore recovers 42\% (Greek) and 40\% (English) of the shallow-to-deep validation gap and 44\% and 41\% of the held-out test gap. This falls within the roughly 34--50\% of the perplexity gap reported by Saunshi et al.~\cite{ref8}, although their fractions are computed on perplexity at a larger scale and ours on mean NLL; we read it as consistent with their pattern at small bilingual scale.

\subsection{Mixers and local components}\label{mixers-and-local-components}

\begin{table*}[t]
\caption{Dense mixer quality and training cost.}\label{tab:layout3}
\small
\begin{tabularx}{\textwidth}{@{}>{\hsize=1.2500\hsize\raggedright\arraybackslash}X>{\hsize=0.4000\hsize\raggedleft\arraybackslash}X>{\hsize=1.2000\hsize\raggedleft\arraybackslash}X>{\hsize=1.2000\hsize\raggedleft\arraybackslash}X>{\hsize=0.9500\hsize\raggedleft\arraybackslash}X@{}}
\toprule
 & & \multicolumn{2}{c}{\textbf{Validation NLL}} & \\
\cmidrule(lr){3-4}
\textbf{Model} & \centering\arraybackslash \textbf{Seeds} & \centering\arraybackslash \textbf{Greek} & \centering\arraybackslash \textbf{English} & \centering\arraybackslash \textbf{Allocated GPU-hours}\\
\midrule
\rowcolor{groupgray}\multicolumn{5}{@{\hspace{4pt}}l}{\textbf{Three-seed comparisons}}\\
GQA12, original mode & 3 & 3.15366 ± 0.00414 & 3.36003 ± 0.00572 & 9.28\\
Dense hybrid 3:1 & 3 & 3.15454 ± 0.01718 & 3.35447 ± 0.00750 & 11.59\\
Dense hybrid 1:1 & 3 & 3.14328 ± 0.01349 & 3.34644 ± 0.01060 & 10.92\\
\rowcolor{groupgray}\multicolumn{5}{@{\hspace{4pt}}l}{\textbf{Single-seed screen}}\\
All-GDN & 1 & 3.25590 & 3.42495 & 12.07\\
\bottomrule
\end{tabularx}
\par\vspace{3pt}{\footnotesize\raggedright\textbf{Notes.} NLL is in nats per target (lower is better); entries show mean ± sample SD. GPU-hours are mean allocated training time. No SD is estimated for the single-seed GDN screen.\par}
\end{table*}

Dense mixer differences are an order of magnitude smaller than the capacity effects (Table 4). The 3:1 hybrid's mean difference from GQA is +0.00088 Greek and −0.00557 English; both language rankings reverse between seed 42 and seeds 43/44, and it costs 2.32 more GPU-hours. Its Greek seed SD (0.01718) is about four times that of GQA, largely from a higher seed 42 endpoint (3.17410). The 1:1 hybrid lowers mean validation loss relative to GQA by 0.01038 Greek and 0.01360 English, improving English at all three paired seeds and Greek at two, at 1.64 more GPU-hours and 4.8M more parameters. On the held-out test it improves both languages at all three paired seeds, which makes it the one dense variant that beats GQA consistently. The single-seed all-GDN endpoint is worse than GQA in both languages. Kimi Linear found 3:1 best at equal FLOPs with 1:1 close behind \cite{ref3}; our equal-target result favors 1:1. Mixer families, budget and scale differ, so the results are not in direct conflict, but they caution against treating 3:1 as a universal default.

\begin{table*}[t]
\caption{Exploratory component screens at seed 42.}\label{tab:layout6}
\small
\begin{tabularx}{\textwidth}{@{}>{\hsize=1.2400\hsize\raggedright\arraybackslash}X>{\hsize=0.6400\hsize\raggedleft\arraybackslash}X>{\hsize=0.6400\hsize\raggedleft\arraybackslash}X>{\hsize=1.4800\hsize\raggedright\arraybackslash}X@{}}
\toprule
 & \multicolumn{2}{c}{\textbf{Validation NLL}} & \\
\cmidrule(lr){2-3}
\textbf{Intervention} & \centering\arraybackslash \textbf{Greek} & \centering\arraybackslash \textbf{English} & \textbf{Comparison}\\
\midrule
\rowcolor{groupgray}\multicolumn{4}{@{\hspace{4pt}}l}{\textbf{Positional encoding}}\\
Full-RoPE GQA baseline & 3.15523 & 3.36063 & Original numerical mode\\
Half-dimensional RoPE & 3.17318 & 3.38395 & Other GQA controls held fixed\\
No positional rotation & 3.25649 & 3.49349 & Other GQA controls held fixed\\
\rowcolor{groupgray}\multicolumn{4}{@{\hspace{4pt}}l}{\textbf{Local convolution and attention placement}}\\
Hybrid convolution width 2 & 3.16853 & 3.36632 & Width 4 baseline: 3.17410 / 3.36114\\
Hybrid attention moved to final three layers & 3.19886 & 3.39660 & Same mixer counts as periodic 3:1 baseline\\
\rowcolor{groupgray}\multicolumn{4}{@{\hspace{4pt}}l}{\textbf{Recurrent family substitution}}\\
KDA/GQA MoE16 & 3.06419 & 3.22581 & GDN/GQA MoE16: 3.08120 / 3.25138\\
\bottomrule
\end{tabularx}
\par\vspace{3pt}{\footnotesize\raggedright\textbf{Notes.} All screens use seed 42. NLL is in nats per target (lower is better); paired baseline losses in the comparison column are Greek / English. These results are exploratory.\par}
\end{table*}

Table 5 lists the single-seed component screens. As a yardstick, the three-seed dense contrasts have paired-difference SDs of 0.012--0.016 (Greek) and 0.004--0.009 (English). Against that yardstick, removing rotation (+0.101 / +0.133 nats) and moving attention to the tail (+0.025 / +0.035) are larger than seed noise in both languages; half-dimensional rotation (+0.018 / +0.023) is at the edge of the Greek range but above the English range; and the width-2 convolution (−0.006 / +0.005) is within it.

\begin{table*}[t]
\caption{Paired KDA validation results.}\label{tab:layout7}
\small
\begin{tabularx}{\textwidth}{@{}>{\hsize=0.5000\hsize\raggedright\arraybackslash}X>{\hsize=1.1250\hsize\raggedleft\arraybackslash}X>{\hsize=1.1250\hsize\raggedleft\arraybackslash}X>{\hsize=1.1250\hsize\raggedleft\arraybackslash}X>{\hsize=1.1250\hsize\raggedleft\arraybackslash}X@{}}
\toprule
 & \multicolumn{2}{c}{\textbf{KDA validation NLL}} & \multicolumn{2}{c}{\textbf{Delta versus GDN}}\\
\cmidrule(lr){2-3}\cmidrule(l){4-5}
\textbf{Seed} & \centering\arraybackslash \textbf{Greek} & \centering\arraybackslash \textbf{English} & \centering\arraybackslash \textbf{Greek} & \centering\arraybackslash \textbf{English}\\
\midrule
42 & 3.06419 & 3.22581 & −0.01700 & −0.02557\\
43 & 3.04426 & 3.21707 & −0.01994 & −0.01380\\
44 & 3.04800 & 3.21754 & −0.02327 & −0.02082\\
\bottomrule
\end{tabularx}
\par\vspace{3pt}{\footnotesize\raggedright\textbf{Notes.} Delta = KDA minus GDN NLL at the same seed; negative favors KDA. Values are in nats per target. All three paired seeds are complete. Across three paired seeds, mean KDA NLL is 3.05215 ± 0.01060 Greek and 3.22014 ± 0.00492 English. Mean paired differences are −0.02007 nats per target in both languages.\par}
\end{table*}

The KDA variant improves on GDN/GQA MoE16 in both languages at every paired seed (Table 6), the largest three-seed mixer effect in the study. Attribution to the recurrence is unresolved: replacing the inherited SiLU output gate with a sigmoid output gate (not the source method's low-rank form) changes validation loss by −0.026 (Greek) and −0.024 (English) in a two-seed supplementary screen (Appendix E.4), an effect as large as the KDA-versus-GDN difference itself, and no GDN arm with the same sigmoid gate was run.

\subsection{Held-out test, bits per character and Greek components}\label{held-out-test-bits-per-character-and-greek-components}

The held-out test (Table 12, Table 14, Figure 3) preserves the direction of the expert-capacity, shared-versus-shallow, shared-versus-independent-depth and KDA comparisons at every paired seed in both languages. MoE16 versus dense 3:1 differs by −0.07892 Greek and −0.11249 English, MoE32 versus MoE16 by −0.02225 and −0.03082, and 1:1 versus GQA by −0.01014 and −0.01341 at all three seeds. The paired document-bootstrap 95\% intervals exclude zero for every contrast. For the small dense-mixing contrasts this understates total uncertainty: in Greek, the seed SD of the 3:1-versus-GQA and 1:1-versus-3:1 paired differences (0.0148 and 0.0116) is several times the bootstrap half-width (about 0.0016).

\begin{figure*}[t]
\centering
\includegraphics[width=1.0\textwidth]{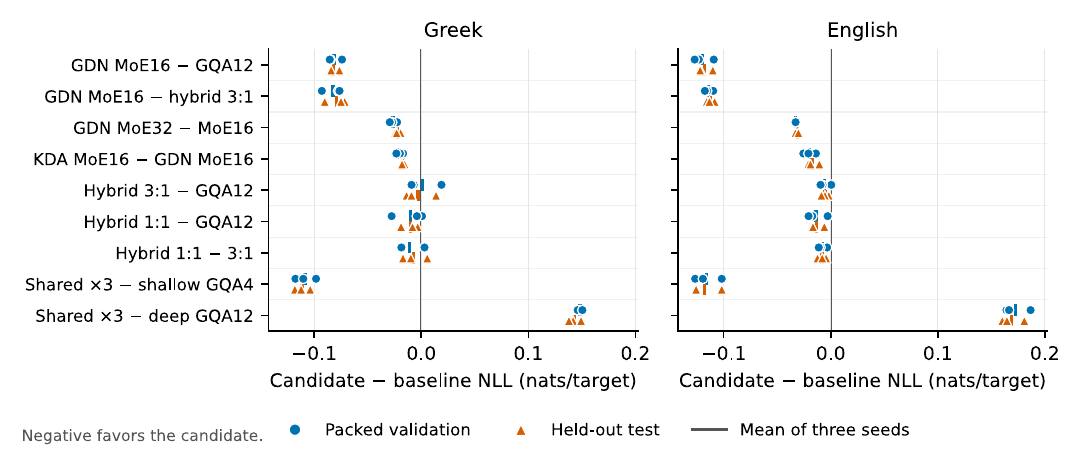}
\caption{Candidate-minus-baseline NLL, in nats per target, for all nine paired contrasts at each of the three paired seeds, on packed validation (circles) and the document-segmented held-out test (triangles). Short bars mark the three-seed means; negative values favor the candidate. Both language panels share one scale. GQA12 family comparisons use original mode; the shared-versus-deep contrast uses the matched-mode GQA control. No confidence interval or equal-compute comparison is implied.}\label{fig:3}
\end{figure*}

\textbf{Tokenizer fertility.} The shared 32,000-token tokenizer is near parity between the languages: held-out Greek text averages 3.92 characters per token and English 4.17, so Greek needs about 6\% more tokens per character (ratio 1.063), and 1.62 versus 1.47 tokens per whitespace word. Within Greek, web text (4.28 characters per token) tokenizes most efficiently and synthetic mathematics least (3.39).

\textbf{One scale for both languages.} Per-target NLL makes English look harder (3.28854 versus 3.05028 nats for original-mode GQA), but per character the two languages are close. Greek BPC is 1.123 for GQA and 1.085 for MoE32, against English 1.138 and 1.087; for every three-seed group Greek is 0.002--0.021 bits per character below English (differences of seed means). English is predicted about as well despite its much smaller training share. The test pools differ in content, so this does not measure relative language difficulty.

\textbf{Greek components.} Synthetic mathematics, drawn from a collection outside the training sources, is the hardest Greek component per character (1.425 BPC for GQA against 0.984 for stories, 0.991 for web and 1.259 for reference text) and gains least from routed experts. From GQA to MoE32, BPC falls by 1.2\% on synthetic mathematics against 4.2--5.1\% on the other components (relative differences of seed means), and in NLL, MoE32 versus MoE16 improves synthetic mathematics by 0.009 nats against 0.027--0.032 on the other components (Table 15). Capacity gains are therefore smallest on synthetic mathematics and largest on stories, not concentrated in the synthetic component. Shared-depth contrasts keep their direction in every component at every seed, with synthetic mathematics gaining as much as the others. The KDA gain is largest on synthetic mathematics (−0.027) and smallest on web text (−0.011), and the 3:1-versus-GQA difference changes sign across components.

\subsection{Training stability}\label{training-stability}

All 32 selected runs completed: their logs contain one loss and gradient-norm record for every update from 1 to 366,211 (11,718,752 records), with no missing or duplicated updates and no non-finite loss or gradient norm. Clipping was nearly always active: after warmup, 98.715--99.990\% of logged pre-clipping norms exceed 1, with group medians of 2.6 to 3.6 (Table 16, Figure 5). Because AdamW normalizes update magnitude, a near-constant rescaling need not change updates much. Appendix E.6 gives trajectories and a spike diagnostic.

\section{Cached-inference consistency and cost}\label{cached-inference-consistency-and-cost}

We check cached execution on completed seed 42 checkpoints by comparing BF16 logits from incremental execution with those from full-prefix execution under the same policy. An element passes when \(|f_i-c_i|\le a+r|c_i|\), with absolute/relative tolerances 0.125/0.02 for the GQA/GDN families and 0.04/0.04 for KDA; for a logit of magnitude 20 these span about four and seven BF16 steps. Every final passing suite shows zero observed difference. Tests are restricted to one document, batch size one and contexts up to 2,048 tokens.

\begin{table*}[t]
\caption{Bounded cached-inference consistency checks.}\label{tab:layout8}
\small
\begin{tabularx}{\textwidth}{@{}>{\hsize=1.0800\hsize\raggedright\arraybackslash}X>{\hsize=0.6400\hsize\raggedleft\arraybackslash}X>{\hsize=0.6800\hsize\raggedleft\arraybackslash}X>{\hsize=1.6000\hsize\raggedright\arraybackslash}X@{}}
\toprule
 & \multicolumn{2}{c}{\textbf{Passing / total checks}} & \\
\cmidrule(lr){2-3}
\textbf{Model} & \centering\arraybackslash \textbf{Synthetic} & \centering\arraybackslash \textbf{Real text} & \textbf{Interpretation}\\
\midrule
\rowcolor{groupgray}\multicolumn{4}{@{\hspace{4pt}}l}{\textbf{Attention and shared-depth models}}\\
Deep GQA & 850 / 850 & 520 / 520 & Exact observed logits throughout these suites.\\
Shallow GQA & 850 / 850 & 520 / 520 & Exact observed logits throughout these suites.\\
Looped GQA & 850 / 850 & 520 / 520 & Exact observed logits; separate visit caches required.\\
\rowcolor{groupgray}\multicolumn{4}{@{\hspace{4pt}}l}{\textbf{GDN expert hybrids}}\\
GDN/GQA MoE16 & 850 / 850 & 520 / 520 & Real-text suite also passes with GPU expert dispatch.\\
GDN/GQA MoE32 & 850 / 850 & 520 / 520 & Real-text suite also passes with GPU expert dispatch.\\
\rowcolor{groupgray}\multicolumn{4}{@{\hspace{4pt}}l}{\textbf{KDA hybrid with corrected convolution}}\\
KDA/GQA MoE16, corrected convolution & 850 / 850 & 1,040 / 1,040 & Exact observed logits under a separately evaluated inference policy.\\
\bottomrule
\end{tabularx}
\par\vspace{3pt}{\footnotesize\raggedright\textbf{Notes.} Entries are passing / total comparisons, not accuracy scores. All checkpoints are seed 42. The policies, tolerances and document sets differ as described in the text; KDA requires its separately qualified convolution correction.\par}
\end{table*}

Under the original execution policy, all five initially audited models fail incremental checks despite exact prefill (Table 10, Appendix C). A fixed-schedule BF16 policy that stabilizes general matrix multiplication (GEMM) and normalization reductions and uses fixed-tile FlexAttention makes them pass (Table 7); recurrent caches retain convolution history and replay state, and looped GQA retains a cache per logical visit. KDA passed synthetic checks but then failed two late comparisons on a real Greek document; a separate fixed-order convolution policy with bounded chunk replay passes all 1,890 synthetic and real comparisons, on eight validation documents against four for the other models.

The qualified policies are self-consistent but not identical to native evaluation. On the full 131,072 validation targets per language, maximum absolute mean NLL change is 0.000271 across the five non-KDA models, preserving their ordering, and +0.000089 Greek and +0.000003 English for corrected KDA. Next-token argmax agreement with native execution is 97.19--98.61\% for the five non-KDA models and 97.65\% (Greek) and 98.23\% (English) for corrected KDA, so the policies change the most likely next token at 1.4--2.8\% of positions.

\begin{table*}[t]
\caption{Qualified inference throughput and memory.}\label{tab:layout9}
\small
\begin{tabularx}{\textwidth}{@{}>{\hsize=1.7000\hsize\raggedright\arraybackslash}X>{\hsize=0.7500\hsize\raggedleft\arraybackslash}X>{\hsize=0.8500\hsize\raggedleft\arraybackslash}X>{\hsize=0.7500\hsize\raggedleft\arraybackslash}X>{\hsize=0.9500\hsize\raggedleft\arraybackslash}X@{}}
\toprule
\textbf{Model and qualified policy} & \centering\arraybackslash \textbf{Decode tokens/s (median)} & \centering\arraybackslash \textbf{Range of three repeats} & \centering\arraybackslash \textbf{Retained cache, MiB} & \centering\arraybackslash \textbf{BF16 weights, MiB}\\
\midrule
\rowcolor{groupgray}\multicolumn{5}{@{\hspace{4pt}}l}{\textbf{Attention and shared-depth models}}\\
Deep GQA & 63.45 & 63.38--63.65 & 24.00 & 217.9\\
Shallow GQA & 186.97 & 186.96--187.21 & 8.00 & 103.9\\
Looped GQA, three visits & 63.56 & 63.45--63.63 & 24.00 & 103.9\\
\rowcolor{groupgray}\multicolumn{5}{@{\hspace{4pt}}l}{\textbf{GDN expert hybrids}}\\
GDN/GQA MoE16, CPU expert dispatch & 31.57 & 31.39--31.67 & 9.49 & 799.0\\
GDN/GQA MoE16, GPU expert dispatch & 34.02 & 33.96--34.16 & 9.49 & 799.0\\
GDN/GQA MoE32, CPU expert dispatch & 31.28 & 31.24--31.36 & 9.49 & 1,447.3\\
GDN/GQA MoE32, GPU expert dispatch & 33.83 & 33.70--33.86 & 9.49 & 1,447.3\\
\rowcolor{groupgray}\multicolumn{5}{@{\hspace{4pt}}l}{\textbf{KDA hybrid with corrected convolution}}\\
KDA/GQA MoE16, corrected convolution & 31.70 & 31.67--31.73 & 9.49 & 802.3\\
\bottomrule
\end{tabularx}
\par\vspace{3pt}{\footnotesize\raggedright\textbf{Notes.} Batch size one; 1,984-token prefix plus 64 generated tokens; one warmup and three measured repeats on NVIDIA B200 hardware, eager execution. Cache excludes weights and temporary allocations; weights are unique parameters in BF16. MoE CPU and GPU dispatch rows are paired measurements from one session.\par}
\end{table*}

Table 8 reports decoding at batch size one with a 1,984-token prefix and 64 generated tokens on NVIDIA B200 hardware. Decoding runs eagerly, without CUDA graphs or compilation, so per-token time is dominated by launch and dispatch overhead roughly proportional to executed layers: shallow GQA decodes 2.95 times faster than deep GQA, and looped GQA matches deep GQA. GPU expert dispatch raises MoE16 and MoE32 throughput by 1.08 times while preserving the checked outputs. At this context the dominant memory term is weights, not cache: MoE32 needs about 1.42 GiB for weights and cache together, versus 0.24 GiB for deep GQA.

\section{Conditional architectural decisions}\label{conditional-architectural-decisions}

Every row of Table 9 is conditional on the shared, untuned AdamW recipe and on the inference policies of Section 5.

\begin{table*}[t]
\caption{Conditional architectural decisions.}\label{tab:layout10}
\small
\begin{tabularx}{\textwidth}{@{}>{\hsize=0.5400\hsize\raggedright\arraybackslash}X>{\hsize=1.4600\hsize\raggedright\arraybackslash}X@{}}
\toprule
\textbf{Constraint or objective} & \textbf{Evidence and boundary}\\
\midrule
\rowcolor{groupgray}\multicolumn{2}{@{\hspace{4pt}}l}{\textbf{Resource constraints}}\\
Dense model at the lowest observed training allocation & \textbf{Evidence:} Among the 12-layer dense families, GQA has the lowest allocation. The 1:1 hybrid lowers loss by about 0.010--0.014 nats at 1.64 more GPU-hours (18\% more), whereas the 3:1 hybrid shows no consistent difference at 2.32 more GPU-hours.\par \textbf{Boundary:} Allocations are device-hours under one recipe; this is not a complete equal-GPU-hour frontier or a latency ranking.\\
More stored expert capacity is affordable & \textbf{Evidence:} Routed experts give the largest gains at matched approximate active parameters, and thirty-two experts improve both language losses over sixteen at every paired seed with a small observed allocation difference; BF16 weights grow from 799 to 1,447 MiB while retained cache stays at 9.49 MiB.\par \textbf{Boundary:} Additional total capacity is part of the intervention; no equal-memory advantage is established.\\
Unique weight storage is constrained & \textbf{Evidence:} Three shared passes improve over one pass at the same unique parameter count (104 MiB of BF16 weights), recovering about 40--44\% of the gap to independent depth.\par \textbf{Boundary:} Executed computation also triples and no independent model with matched unique parameters was trained; logical cache equals the deep model's.\\
\rowcolor{groupgray}\multicolumn{2}{@{\hspace{4pt}}l}{\textbf{Mixer and component choices}}\\
Quality within a recurrent/attention hybrid is the priority & \textbf{Evidence:} The 1:1 hybrid improves on both the 3:1 hybrid and GQA; against GQA it improves both languages at every held-out seed.\par \textbf{Boundary:} Greek validation reverses at one seed in both contrasts and parameter counts differ; Kimi Linear found 3:1 best at equal FLOPs \cite{ref3}.\\
A positional or local component must be selected & \textbf{Evidence:} In seed 42 screens, removing rotation or moving attention to the tail worsens loss by more than seed noise; half rotation is at the edge of the Greek noise range but above the English one, and width-2 convolution is within noise.\par \textbf{Boundary:} Single-seed results do not establish confirmed effects, other context lengths or a universal default.\\
\rowcolor{groupgray}\multicolumn{2}{@{\hspace{4pt}}l}{\textbf{KDA family comparison}}\\
KDA is considered within the tested hybrid recipe & \textbf{Evidence:} All three paired validation seeds and all three held-out seeds improve both languages versus GDN/MoE16; keep this quality comparison separate from inference readiness.\par \textbf{Boundary:} The sigmoid-gate change is as large as the KDA gain, so attribution to the recurrence is unresolved; the corrected convolution policy passes its bounded suite, not general inference correctness.\\
\bottomrule
\end{tabularx}
\par\vspace{3pt}{\footnotesize\raggedright\textbf{Notes.} Recommendations are conditional on the tested scale, data mixture and training recipe. Evidence and limitations are retained together for each decision.\par}
\end{table*}

\section{Limitations}\label{limitations}

\textbf{Scope.} The study uses one width envelope, one bilingual mixture, one tokenizer, one 2k context and a shared optimizer schedule without per-family tuning. Greek--English is the application setting, not an experimental factor, so no result establishes language-specific causation or transfer to other language pairs. Loss differences of 0.01--0.03 nats per target, the size of the mixer, expert-count and KDA effects, correspond to perplexity changes of roughly 1--3\%; whether differences of this size change downstream task behavior is not evaluated in this paper.

\textbf{Statistics and selection.} Three seeds give descriptive evidence from a small sample, not calibrated significance tests; single-seed screens are exploratory. The campaign is adaptive and its questions are organized retrospectively; it is not a preregistered factorial study. Reusing the validation pools for selection may introduce selection bias. The document bootstrap, the Greek component breakdown and the BPC and fertility analyses were added after test scoring and are descriptive.

\textbf{Controls.} Total parameters, exact computation and memory are not matched across families. Expert-count comparisons intentionally increase capacity, and no equal-total-capacity dense control isolates a general sparsity advantage. No independently parameterized model matches the shared-depth model's unique parameters, so its gain over one pass cannot be separated from the threefold increase in executed computation; each shared parameter also accumulates gradient from three visits before the global clip, which we do not analyze.

\textbf{Compute.} All comparisons are at equal training targets, not equal compute. Comparing at equal elapsed time from the existing validation trajectories is inconclusive, because the more expensive model is mid-schedule, with its cosine decay unfinished, at the baseline's endpoint: for example, MoE32 at about 54\% of its schedule is 0.13--0.17 nats worse than finished GQA12, which reflects the unfinished schedule rather than an equal-compute result. Equal compute remains untested. Training costs were measured on NVIDIA B200 GPUs and are not portable performance figures; decoding rates describe a correctness-first implementation, not architecture-intrinsic speed.

\textbf{Numerics and inference.} Cache qualification covers one document, batch one, contexts up to 2,048 tokens and four or eight validation documents; no model is qualified for general multi-request serving or longer contexts. CUDA recurrence audits against an FP64 oracle retain failed state and gradient comparisons under fixed elementwise tolerances, all with relative L2 error below 1.3\% (Appendix D); their effect on training quality is not measured.

\textbf{Data.} The corpus includes unpublished collections, and training-source provenance, licenses and privacy screening are incomplete. Exact and normalized 13-word overlap exclusion is the only decontamination check; near-duplicate leakage, including into the synthetic Greek test components, was not measured.

\section{Conclusion}\label{conclusion}

At this tiny scale and fixed training-target budget, stored capacity matters far more than mixer design. Routed experts give large gains at every paired seed, and more experts give more; shared depth recovers a fraction of the shallow-to-deep gap similar to that reported for larger looped models but trails independent depth and keeps a cache per logical visit. Mixer choices move loss by about 0.01--0.02 nats, with the 1:1 hybrid the only dense variant that beats GQA at every held-out seed. Per character, Greek and English are predicted about equally well. Cached-inference consistency is a separate requirement: it held only under the documented policies and suites, and those policies change a few percent of next-token choices relative to native evaluation.

\section*{Reproducibility and declarations}\label{reproducibility-and-declarations}
\addcontentsline{toc}{section}{Reproducibility and declarations}

\textbf{Data and code availability.} The training corpus includes unpublished, locally held collections. Data, checkpoints, model and evaluation code, resolved training configurations and per-seed experimental records are not publicly released with this paper. The manuscript reports the architecture (Section 3.1 and Appendix A), the complete training hyperparameters and hardware class (Section 3.3), the evaluation protocols, aggregate and paired results with their dispersion, and limitations. Independent repetition of training and checkpoint evaluation requires access to the withheld materials.

\textbf{Ethics and data limitations.} The work trains models on an existing text corpus and reports no new participant study. This does not establish that the corpus contains no personal or sensitive information. Source provenance, licenses and privacy screening are incomplete; no institutional approval or exemption is asserted.

\textbf{Author contributions.} Ayoub Kirouane is the sole author and is responsible for conceptualization, methodology, software, investigation, data curation, analysis, visualization, manuscript preparation and revision.

\textbf{Competing interests.} The author is affiliated with Sophea AI Lab, KIEFER SA, which supported the work. The author declares no additional competing interests.

\textbf{Funding.} KIEFER SA funded the work and provided computing resources.

\appendix

\section{Model and evaluation specification}\label{model-and-evaluation-specification}

\subsection{Notation, residual blocks and feed-forward computation}\label{notation-residual-blocks-and-feed-forward-computation}

We specify the real-arithmetic computation; kernel rounding and execution policies are addressed separately in Section 3.3 and Section 5. Vectors are columns, \(x_t\in\mathbb{R}^{d}\), with model width \(d=768\). There are \(H_q=6\) query heads and \(H_{kv}=2\) KV heads in GQA, with head dimensions \(d_k=d_v=128\). Recurrent layers have six key/value heads. Parameters are layer-specific unless sharing is stated. Bias-free linear maps are written as matrices acting on vectors; elementwise multiplication is \(\odot\).

For a vector \(z\in\mathbb{R}^{m}\), define bias-free RMS normalization \cite{ref18} with numerical stabilizer and the two gate activations as

\begin{equation}
\begin{gathered}
\mathcal{N}_{\gamma,\epsilon}(z)
=\gamma\odot z\left(m^{-1}\sum_{j=1}^{m}z_j^2+\epsilon\right)^{-1/2}\\
\sigma(z)=\frac{1}{1+e^{-z}}\\
\operatorname{SiLU}(z)=z\odot\sigma(z).
\end{gathered}
\label{eq:1}
\end{equation}

Each decoder layer uses two pre-normalized residual updates:

\begin{equation}
\begin{gathered}
u_t=x_t+\operatorname{Mixer}_{\theta}\bigl(\mathcal{N}_{\gamma_a,10^{-6}}(x_t)\bigr)\\
x'_t=u_t+\operatorname{FFN}_{\phi}\bigl(\mathcal{N}_{\gamma_f,10^{-6}}(u_t)\bigr).
\end{gathered}
\label{eq:2}
\end{equation}

The mixer operates on the causally available sequence, not just the displayed token. Dense FFNs and individual experts use the bias-free SwiGLU form \cite{ref17}:

\begin{equation}
\begin{gathered}
f(z)=W_2\left[\operatorname{SiLU}(W_1z)\odot W_3z\right]\\
W_1,W_3\in\mathbb{R}^{d_{\mathrm{ff}}\times d}\\
W_2\in\mathbb{R}^{d\times d_{\mathrm{ff}}}.
\end{gathered}
\label{eq:3}
\end{equation}

\begin{center}
\includegraphics[width=0.78\linewidth]{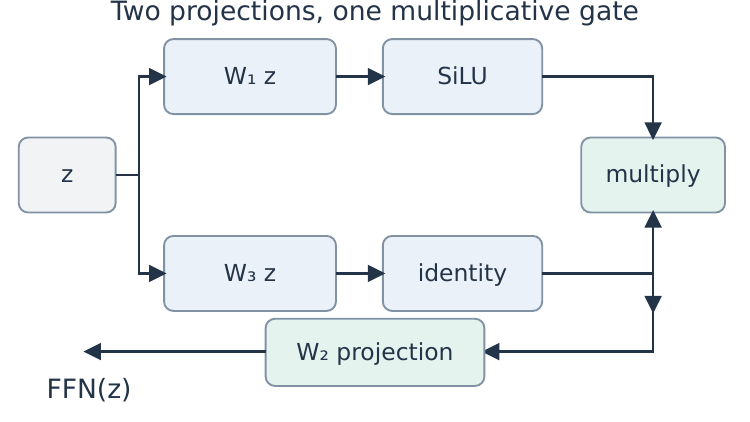}
\captionof{diagram}{SwiGLU computation \cite{ref17}. The two input projections interact through a SiLU gate before the output projection. The surrounding residual updates are defined in Equation 2.}
\end{center}

The dense intermediate width is \(d_{\mathrm{ff}}=2304\); each routed or shared expert uses 768. After all layer executions, one final RMS normalization precedes the vocabulary projection. Input embedding and output projection weights are tied. Multi-token prediction contributes no loss in these base runs.

\subsection{Gated grouped-query attention and position}\label{gated-grouped-query-attention-and-position}

The query projection produces query channels and output-gate channels jointly; denote the latter by \(z_{t,h}\). Before rotary position encoding, each query/key head uses offset RMS normalization, \(\mathcal{N}_{1+\gamma,10^{-6}}\), rather than the direct-gain normalization of Equation 1. Write \(\widehat q_{t,h}\) and \(\widehat k_{t,j}\) for these normalized heads. For an even rotated dimension \(r\), RoPE applies planar rotations with angle \(t\theta^{-2i/r}\), \(0\leq i<r/2\), to coordinate pairs \((i,i+r/2)\) in the first \(r\) channels; the remaining channels are unchanged. Here \(\theta=500000\), and \(r=128\) or 64 for full/half RoPE. NoPE uses the identity transform (\(r=0\)), without evaluating the frequency formula. Positions reset at document boundaries. The block uses the same token positions on each loop visit.

Let \(q_{t,h}=R_r(t)\widehat q_{t,h}\), \(k_{t,j}=R_r(t)\widehat k_{t,j}\) and \(j(h)=\lfloor h/(H_q/H_{kv})\rfloor\) for zero-based head indices. For tokens \(s\) allowed by the causal, same-document mask \(\mathcal{A}(t)\),

\begin{equation}
\begin{gathered}
a_{t,s,h}=\frac{\exp(q_{t,h}^{\top}k_{s,j(h)}/\sqrt{d_k})}
{\sum_{u\in\mathcal{A}(t)}\exp(q_{t,h}^{\top}k_{u,j(h)}/\sqrt{d_k})}\\
b_{t,h}=\sum_{s\in\mathcal{A}(t)}a_{t,s,h}v_{s,j(h)},
\end{gathered}
\label{eq:4}
\end{equation}

\begin{equation}
\operatorname{GQA}(x)_t
=W_o\operatorname{concat}_{h}\left[b_{t,h}\odot\sigma(z_{t,h})\right].
\label{eq:5}
\end{equation}

\begin{center}
\includegraphics[width=0.78\linewidth]{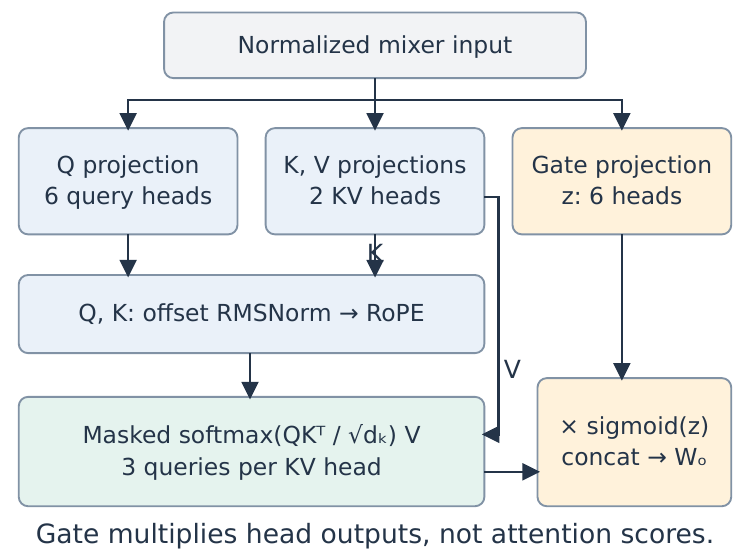}
\captionof{diagram}{Study gated-GQA computation. KV sharing follows GQA \cite{ref1}, the per-head sigmoid output gate follows \cite{ref20}, and the offset (zero-centered) normalization matches the Qwen3-Next design \cite{ref21}. Six query heads share two KV heads. Query and gate channels come from one joint projection, shown as separate paths. Only queries and keys receive offset normalization and positional rotation; the sigmoid gate acts on attention outputs before the output projection.}
\end{center}

The gate acts on attention outputs before \(W_o\), not on attention scores. Padding is excluded from valid targets; attention metadata enforces document isolation.

\subsection{Local convolution, GDN and KDA}\label{local-convolution-gdn-and-kda}

Both recurrent families first project the normalized layer input into query, key and value streams. Each stream receives a bias-free depthwise causal convolution and SiLU. For projected channel sequence \(p\),

\begin{equation}
c_{t,j}=\operatorname{SiLU}\left(\sum_{i=0}^{w-1}a_{j,i}p_{t-w+1+i,j}\right).
\label{eq:6}
\end{equation}

Weights are indexed from the oldest to newest tap; missing history is zero at each document start. The baseline width is \(w=4\) and the local-width screen uses 2. Prepared queries and keys use \(\operatorname{L2}(c)=c/\sqrt{\sum_jc_j^2+10^{-6}}\) per head; values use the convolved stream without this normalization.

Use state \(S_t\in\mathbb{R}^{d_k\times d_v}\), initially zero per document. Denote prepared vectors by \(q_t,k_t,v_t\) and the write gate by \(\beta_t\in(0,1)\). A common recurrence is

\begin{equation}
\begin{gathered}
\overline S_t=D_tS_{t-1}\\
r_t=v_t-\overline S_t^{\top}k_t\\
S_t=\overline S_t+\beta_t k_t r_t^{\top}\\
o_t=S_t^{\top}\frac{q_t}{\sqrt{d_k}}.
\end{gathered}
\label{eq:7}
\end{equation}

\begin{center}
\includegraphics[width=0.78\linewidth]{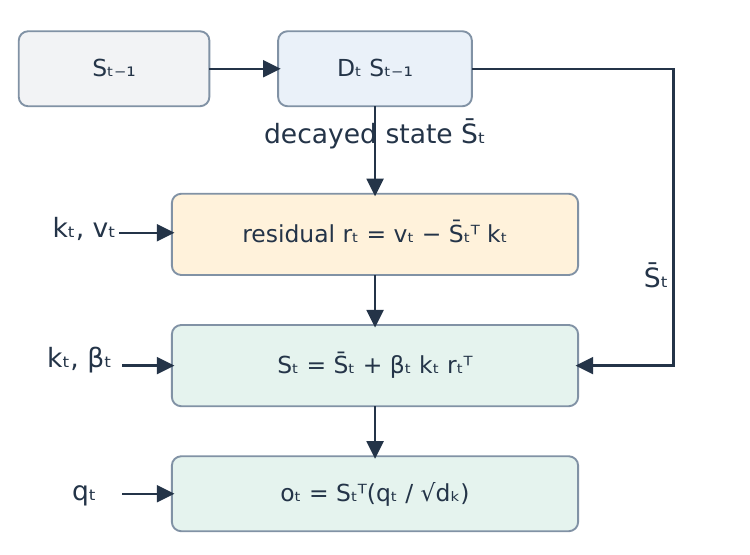}
\captionof{diagram}{Delta recurrence within one head. Decay is applied before the residual write, followed by the query readout. The state is zero at each document start. Output normalization and gating follow Equation 10.}
\end{center}

Equivalently, \(S_t=(I-\beta_t k_tk_t^{\top})D_tS_{t-1}+\beta_t k_tv_t^{\top}\). Decay precedes the delta update. These matrices generally do not commute when decay is channelwise.

For scalar-gated GDN \cite{ref2}, each head uses projected scalars \(a_t,b_t\), learned scalar \(A\) and bias \(\delta\):

\begin{equation}
\begin{gathered}
g_t=-e^{A}\operatorname{softplus}(a_t+\delta)\\
D_t=e^{g_t}I\\
\beta_t=\sigma(b_t).
\end{gathered}
\label{eq:8}
\end{equation}

Our KDA-based variant adopts the channelwise recurrence \cite{ref3} and projects forget inputs using two linear maps, \(f_t=W_bW_ax_t\), with intermediate rank 128; the implementation has no nonlinearity between these maps. The following bounded log-decay specifies our installed operator; the source paper describes the decay function more generally. For each head, \(A\) is scalar and \(\delta_j\) is channelwise:

\begin{equation}
\begin{gathered}
g_{t,j}=-5\sigma\!\left(e^{A}(f_{t,j}+\delta_j)\right)\\
D_t=\operatorname{diag}(e^{g_{t,1}},\ldots,e^{g_{t,d_k}})\\
\beta_t=\sigma(b_t).
\end{gathered}
\label{eq:9}
\end{equation}

\begin{center}
\includegraphics[width=0.78\linewidth]{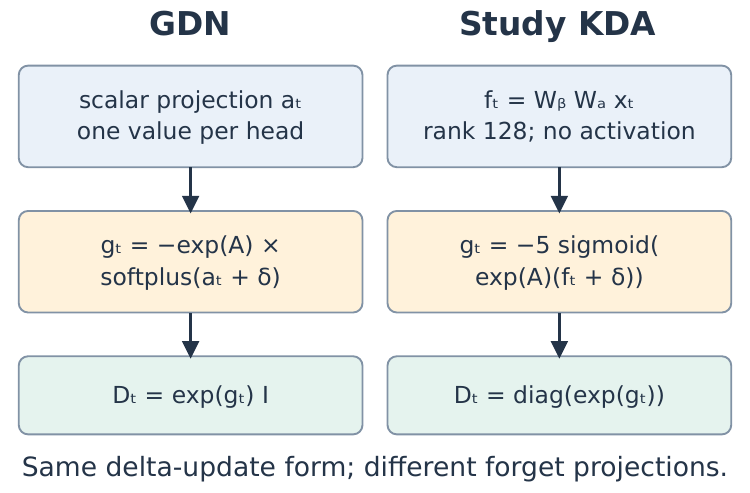}
\captionof{diagram}{Decay parameterization. GDN uses one scalar decay per head; the study KDA variant uses one decay per key channel. Both feed the ordered update in Equation 7. The output gate, omitted here, remains SiLU in both study families.}
\end{center}

Here \(x_t\) in the projection formulas denotes the mixer input after block normalization. The gate \(g\) is a \textbf{natural-log decay}: the public operator consumes \(g\), while low-level log2 routines receive \(g/\ln2\). Query scaling in Equation 7 occurs once. Scalar diagonal decay recovers the GDN recurrence algebra, but not the complete GDN model: the forget projection and gate parameterization still differ.

The recurrent output is normalized per head, gated and projected:

\begin{equation}
\operatorname{Rec}(x)_t=W_o\operatorname{concat}_{h}
\left[\mathcal{N}_{\gamma_o,\epsilon_o}(o_{t,h})\odot\psi(z_{t,h})\right],
\label{eq:10}
\end{equation}

where \(z\) is a learned projection of the mixer input. Both study families use \(\psi=\operatorname{SiLU}\) and \(\epsilon_o=10^{-6}\). The study builder replaces the upstream KDA output norm/gate with the GDN configuration. This also differs from the source paper's low-rank sigmoid output gate: its Section 5.2 ablation favors sigmoid over Swish \cite{ref3}, making our inherited gate a material adaptation rather than an exact reproduction. The forget projection and decay parameterization still make this a family substitution, not a pure scalar-versus-vector decay ablation.

\subsection{Expert routing and balancing}\label{expert-routing-and-balancing}

Following the sigmoid-affinity routing of DeepSeek-V3 \cite{ref16}, the auxiliary-loss-free selection bias of \cite{ref22} and the shared expert of \cite{ref23}, for an MoE FFN input \(x\), router scores are \(s_i=\sigma((W_rx)_i)\). Let \(b_i^{\mathrm{route}}\) be the non-gradient balancing bias. Selection uses biased scores, while mixture weights use the original scores:

\begin{equation}
\begin{aligned}
\mathcal{I}(x)&=\operatorname{TopK}_{i}(s_i+b_i^{\mathrm{route}},2),\\
\omega_i(x)&=\frac{s_i}{\sum_{j\in\mathcal{I}(x)}s_j+10^{-20}},\\
\operatorname{MoE}(x)&=f_{\mathrm{shared}}(x)+\sum_{i\in\mathcal{I}(x)}\omega_i(x)f_i(x). 
\end{aligned}
\label{eq:11}
\end{equation}

There are \(E=16\) or 32 routed experts and one always-active shared expert. Route scale is 1. The discrete selection indices are not differentiated; gradients flow through selected weights and expert computations. In the inspected sign-based balancing hook, counts \(n_i\) exclude padding and accumulate assignments over the optimizer interval. With \(\overline n=E^{-1}\sum_i n_i\) and \(\eta_b=0.001\),

\begin{equation}
\begin{gathered}
\Delta b_i=\eta_b\operatorname{sign}(\overline n-n_i)\\
b_i^{\mathrm{route}}\leftarrow b_i^{\mathrm{route}}+\Delta b_i-E^{-1}\sum_j\Delta b_j.
\end{gathered}
\label{eq:12}
\end{equation}

\begin{center}
\includegraphics[width=0.78\linewidth]{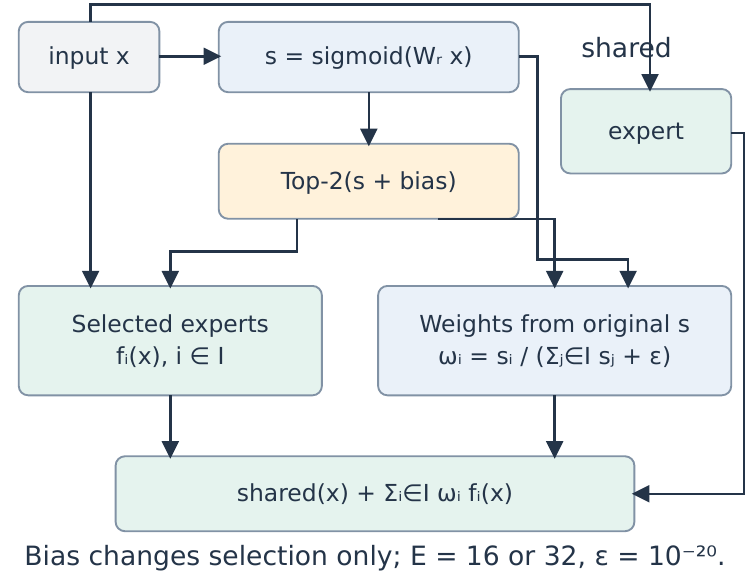}
\captionof{diagram}{Study MoE computation, using the routing of \cite{ref16}, the selection bias of \cite{ref22} and the shared expert of \cite{ref23}. Biased scores choose two routed experts; their original scores determine mixture weights. The shared expert is always active. Equation 12 updates the selection bias between optimizer intervals.}
\end{center}

Counts are then reset. This is a selection-bias update, not an additional loss term. Recorded CPU router checks cover selection, weighting, conditional gradients and masked counts.

\subsection{Shared depth and logical cache state}\label{shared-depth-and-logical-cache-state}

Let \(F_\theta\) denote the composition of the four registered GQA/dense layers. With token embeddings \(h^{(0)}\), the looped model computes

\begin{equation}
\begin{gathered}
h^{(r+1)}=F_\theta(h^{(r)};\mathrm{positions})\\
r=0,1,2\\
\mathrm{logits}=W_{\mathrm{vocab}}\mathcal{N}_{\gamma,10^{-6}}(h^{(3)}).
\end{gathered}
\label{eq:13}
\end{equation}

All block parameters, including norms, are shared; there is no loop embedding or intermediate vocabulary head. Unlike the Universal Transformer \cite{ref7}, this model has no step-coordinate embedding or adaptive halting. Backpropagation sums the chain-rule contributions from all uses of each parameter, while the optimizer owns it once. The shallow control uses one visit; the deep control has twelve independent layers.

\par\medskip\noindent\begin{minipage}{\linewidth}\setlength{\parindent}{1em}
\textbf{Algorithm 1. Shared-block execution with logical caches.} Full training uses the same loop without caches. In incremental execution, cache indexing includes the visit because hidden-state histories differ between visits.

\begin{Verbatim}[fontsize=\scriptsize,frame=single,framesep=3pt,breaklines=true]
h ← Embed(input tokens)
for visit = 0, …, R−1:
    for layer = 0, …, L_unique−1:
        c ← cache[visit, layer]
        h, c ← Block[layer](h, positions, c)
        cache[visit, layer] ← c
return VocabularyProjection(FinalNorm(h))
\end{Verbatim}
\end{minipage}\par\medskip

For full GQA with retained context \(T\) and batch size \(B\), KV storage is

\begin{equation}
M_{\mathrm{KV}}=B L_{\mathrm{logical}}T H_{kv}(d_k+d_v)b_{\mathrm{elem}}.
\label{eq:14}
\end{equation}

\begin{center}
\includegraphics[width=0.78\linewidth]{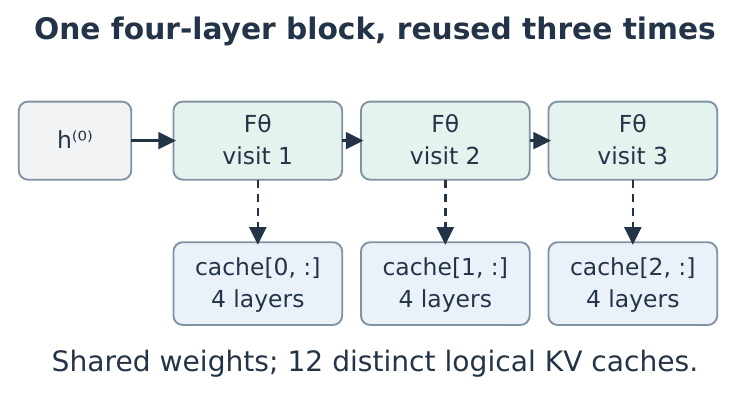}
\captionof{diagram}{Shared-block execution following the looping construction in \cite{ref8}. The three visits use the same four layers and the same token positions. Dashed connections denote separate logical cache ownership during incremental execution, not additional trainable weights.}
\end{center}

With BF16 (\(b_{\mathrm{elem}}=2\) bytes), \(B=1\) and \(T=2048\), twelve logical layers retain 24 MiB; four retain 8 MiB. Consequently the four-layer, three-visit model saves unique weights but retains twelve logical KV caches. Equation 14 excludes recurrent state and convolution history, so it is not total device memory.

\subsection{Objective, evaluation algorithm and verification boundary}\label{objective-evaluation-algorithm-and-verification-boundary}

For valid prediction positions \(\mathcal{V}\), the language-model objective and reported perplexity (PPL) are

\begin{equation}
\begin{gathered}
\mathcal{L}=-\frac{1}{|\mathcal{V}|}\sum_{t\in\mathcal{V}}\log p_\theta(y_t\mid x_{\leq t})\\
\operatorname{PPL}=e^{\mathcal{L}}.
\end{gathered}
\label{eq:15}
\end{equation}

Padding and cross-document transitions are excluded. Validation tables retain the original packed protocol. The completed final evaluation instead scores every eligible within-document next-token pair in nonoverlapping segments of at most 2,048 pairs, resetting positions per segment. This changes available context.

\par\medskip\noindent\begin{minipage}{\linewidth}\setlength{\parindent}{1em}
\textbf{Algorithm 2. Document-segmented held-out evaluation.} Applied separately to the Greek and English manifests for each of the 32 selected endpoints.

\begin{Verbatim}[fontsize=\scriptsize,frame=single,framesep=3pt,breaklines=true]
loss_sum ← float64(0); target_count ← 0
for document tokens z[0:n] in manifest order:
    doc_loss ← float64(0)
    for start = 0, 2048, … while start < n−1:
        end ← min(start+2048, n−1)
        m ← end−start
        inputs ← z[start:end]
        labels ← z[start+1:end+1]
        positions ← 0, …, m−1
        hidden ← NativeHiddenForward(inputs,
                     positions, reset_state=true)
        for offset = 0, 128, … while offset < m:
            stop ← min(offset+128, m)
            states ← hidden[offset:stop]
            targets ← labels[offset:stop]
            logits ← VocabularyProjection(states)
            losses ← CE_FP32(logits, targets)
            doc_loss ← doc_loss + SumFP64(losses)
        target_count ← target_count + m
    loss_sum ← loss_sum + doc_loss
assert target_count = expected_targets(manifest)
return loss_sum / target_count
\end{Verbatim}
\end{minipage}\par\medskip

The model runs in evaluation mode with BF16 weights and no cache reuse across segments. Vocabulary projection is evaluated in groups of 128 positions to limit temporary logits storage; this does not reset attention or split the segment's hidden-state computation. Each retained target contributes once. CE\_FP32 returns unreduced per-target cross-entropy losses. Losses are computed in FP32 and accumulated in FP64, including the final sum across document records.

The equations specify the model, not a proof of all kernels; Appendix D audits the recurrent kernels.

\section{Data and evidence}\label{data-and-evidence}

\textbf{Training data.} The sampled training mixture combines Greek and English web, reference and document text, including unpublished collections held locally. Preparation normalizes text, removes canonical duplicates and excludes exact and normalized 13-word overlap with protected holdouts; these checks do not establish exhaustive semantic or translated-document decontamination. Tokenization appends EOS, and packing excludes cross-document targets. Largest-remainder quotas select complete 2,048-target rows; seeded selection draws without row repetition from each source's prepared supply, followed by shuffling. Upstream revisions and generation lineage are incomplete for some inherited material.

\textbf{Validation and held-out pools.} Several Greek source labels alias the same validation pool and are not independent domain measurements. The collector verifies a unique final endpoint, the expected number of validation targets and identical packed validation data across runs. The held-out builder groups available family identifiers and heuristic near-duplicate links before assigning groups to training, validation and test partitions. Missing family metadata and heuristic grouping limit the independence guarantee. Greek stories and synthetic mathematics together supply 51.44\% of Greek test targets, so the Greek aggregate is a mixture-specific measure, not a natural-web-only evaluation, and neither component is among the training sources; complete generation lineage for those components is not established.

\textbf{Endpoint and evaluator qualification.} The final evaluation includes all 32 predetermined base endpoints. Both later KDA confirmations passed checkpoint integrity, terminal-counter, data-lineage and strict restoration checks before entry. All 64 endpoint/manifest authorizations passed before scoring, and all 32 results passed journal identity, document/target accounting and loss-sum consistency checks before aggregation. The full-validation-pool recovery qualification scored 2,546 documents and 2,095,497 targets, with exact per-document agreement between uninterrupted execution and a fresh-process resume after 17 documents. All eight packed training artifacts passed integrity checks. Across 14 representative configuration/numerical-mode groups, 126 bounded evaluator checks covered restoration, repeatability and accounting. Requested deterministic mode did not establish bitwise trajectory identity; a stronger replay gate failed.

\section{Inference audit}\label{inference-audit}

This appendix summarizes, in order, the controls, failures and measurements behind Section 5. All checks use completed seed 42 checkpoints, single documents, batch one and contexts at most 2,048.

\subsection{Original-policy failures}\label{original-policy-failures}

We first evaluated matched deep GQA, shallow GQA, shared-depth GQA, hybrid MoE16 and hybrid MoE32. Each model underwent 70 cached/full comparisons: two synthetic token patterns, prefix lengths 63, 64, 65, 128, 512, 1,024 and 1,984, and the prefill output followed by four teacher-forced incremental outputs, at BF16 absolute/relative tolerances 0.125/0.02. The recorded implementation calls \texttt{allclose(full,\ cached)}, so the cached operand supplies the relative scale; KDA retains its separately declared pilot thresholds a = r = 0.04. Failed comparisons are not rescued by greedy agreement or by changing the threshold.

\begin{table*}[tb]
\caption{Original-policy cached-logit verification.}\label{tab:layout11}
\small
\begin{tabularx}{\textwidth}{@{}>{\hsize=1.1200\hsize\raggedright\arraybackslash}X>{\hsize=0.9600\hsize\raggedleft\arraybackslash}X>{\hsize=0.9600\hsize\raggedleft\arraybackslash}X>{\hsize=0.9600\hsize\raggedleft\arraybackslash}X@{}}
\toprule
\textbf{Checkpoint} & \centering\arraybackslash \textbf{Within tolerance} & \centering\arraybackslash \textbf{Max. logit error} & \centering\arraybackslash \textbf{Argmax agreement}\\
\midrule
\rowcolor{groupgray}\multicolumn{4}{@{\hspace{4pt}}l}{\textbf{Attention and shared-depth models}}\\
Deep GQA & 46 / 70 & 0.87500 & 70 / 70\\
Shallow GQA & 57 / 70 & 0.90625 & 70 / 70\\
Shared-depth GQA & 39 / 70 & 1.46875 & 70 / 70\\
\rowcolor{groupgray}\multicolumn{4}{@{\hspace{4pt}}l}{\textbf{Expert hybrids}}\\
Hybrid MoE16 & 53 / 70 & 3.62500 & 68 / 70\\
Hybrid MoE32 & 39 / 70 & 4.00781 & 69 / 70\\
\bottomrule
\end{tabularx}
\par\vspace{3pt}{\footnotesize\raggedright\textbf{Notes.} These are original-policy checks, before the corrections described in the text. Fractions are passing / total comparisons; argmax agreement does not establish cached-logit equivalence.\par}
\end{table*}

All prefill comparisons match exactly, yet every model fails at least one incremental comparison, and the expert models also disagree on some greedy decisions (Table 10). No timing was taken under this policy. An earlier KDA pilot had passed its bounded BF16 cache suite after fused replay and native GQA metadata corrections, but the final-checkpoint audit with the same 70-case suite and 0.04 tolerances passes only 44 checks, with maximum absolute logit error 4.1875. We retain both results.

\subsection{Localization}\label{localization}

Tracing the first failing and largest-error cases of the five models (ten fixtures), identical repeats agree exactly, and the first token-aligned difference appears at GQA attention: layer 0 for the dense models and layer 3 for the expert hybrids, with matching current-token query/key/value inputs. Complete historical keys and values at that boundary match exactly in all ten cases, yet native attention produces different outputs for a full query sequence versus one query. For the four MoE cases, one-call and incremental replay of identical projected inputs agree exactly on all nine GDN layers' boundary state, convolution history, pending chunk inputs and final output. Some later expert selections differ: forcing full-prefix expert IDs while retaining cached router scores reduces the worst traced error from 3.625 to 0.25 (MoE16) and from 4.00781 to 0.28125 (MoE32), which supports selection amplification in those cases.

Partial remedies did not generalize. Forcing one attention split lets MoE32 pass 70/70, but an extension to new random seeds and a new bilingual prompt with 64 incremental tokens yields 771/780 despite 780/780 greedy agreement; dense controls that pad the query length, alone or with one split, also fail. FP64 attention with FP32 remaining dense computation passes 70/70 for each dense model (maximum logit difference below 0.0006), but FP64 attention with BF16 remaining MoE computation fails longer held-out tests: 745/780 for MoE16 and 775/780 for MoE32.

\subsection{Fixed-schedule policy}\label{fixed-schedule-policy}

Following the batch-invariance methods of Horace He and collaborators \cite{ref19}, we implemented a process-local inference policy with fixed matrix-multiplication reductions, normalization reductions, per-expert GEMMs and fixed-tile FlexAttention. The policy changes no training dependency or parameter value, and weights stay in BF16. All five models pass 70/70 original checks plus 780/780 extended checks each, with zero observed logit difference in all 4,250 comparisons; the extension uses unseen synthetic prompts, prefixes 63/128/512/1,984 and 64 incremental tokens, reaching context 2,048. A first native/policy bridge on 16,384 held-out targets per language/model gave maximum absolute mean NLL change 0.000870 and argmax agreement 96.94--98.50\%. Extended to the full 131,072 validation targets with identical document segmentation, the maximum change is 0.000271 and agreement 97.19--98.61\%, and both languages retain the ordering MoE32, MoE16, deep GQA, looped GQA, shallow GQA from lowest to highest loss.

The first timings under this policy (median of three repeats after warmup) were 63.45 tokens/s for deep GQA, 186.97 for shallow GQA, 63.56 for looped GQA, 31.46 for MoE16 and 31.36 for MoE32, with retained cache 24 MiB, 8 MiB and 9.49 MiB as in Table 8. A later inference-only optimization removes CPU offset synchronization and per-expert GEMM launches while preserving the fixed reduction tiles. It passes 28 exact kernel checks, 780/780 extended cache comparisons for each GDN/MoE checkpoint with zero observed logit error, and 20 exact full-prefix comparisons against the previous policy. Alternating implementation order within one process, MoE16 rises from 31.57 to 34.02 tokens/s (1.077×) and MoE32 from 31.28 to 33.83 tokens/s (1.081×).

\subsection{KDA}\label{kda}

A first KDA policy, retaining fused paged KDA replay with stable arithmetic around it, passed 850/850 short and extended cache comparisons with zero observed difference at 0.04/0.04 and measured 27.41 tokens/s with 21.33 MiB of retained cache; its full-budget bridge changed mean NLL by +0.000009 Greek and +0.000161 English, with argmax agreement 97.79\% and 98.32\%. A real-document extension then used two Greek and two English 2,048-token validation rows selected by position metadata only, with prefixes 128/1,984 and 64 teacher-forced increments. Deep, shallow and looped GQA and both GDN/MoE variants pass 520/520 each (2,600 exact comparisons), but KDA with GPU dispatch passes only 518/520. Both failures occur near the end of one Greek document, at total lengths 2046 and 2,048, with maximum logit errors 0.125 and 0.25; greedy choices agree. A controlled comparison reproduces both failures with the old and new dispatch at all 65 points of the failing sequence.

A same-input stage audit identifies a nonidentical convolved value at layer 9, length 2043, while convolution history and the normalized inputs at that position agree. Replay-only and packed-metadata-only corrections both retain the two failures. Fixed-order FP32 convolution accumulation in both full and cached execution, combined with bounded ordinary 64-token chunk replay, passes 1,890/1,890 checks exactly, including 1,040 comparisons on eight real validation documents; its bridge changes mean NLL by +0.000089 Greek and +0.000003 English. The earlier KDA policy retains its two real-document failures.

\section{CUDA numerical accuracy}\label{cuda-numerical-accuracy}

CPU FP64 references pass, except 28 strict comparisons of the installed chunk wrapper's FP32 gate accumulation. The CUDA audit compares fused and cuDNN recurrence backends with an independent FP64 sequential oracle on exactly quantized inputs (batch one, six heads, key/value dimension 128, BF16 queries/keys/values, FP32 log decay and beta, lengths 1--2,048 including chunk boundaries, beta limits, strong/zero decay and document resets). Comparisons cover output, final state and five input gradients; eight additional KDA checks cover bounded-gate outputs and derivatives.

\begin{table*}[tb]
\caption{Bounded CUDA numerical verification.}\label{tab:layout12}
\small
\begin{tabularx}{\textwidth}{@{}>{\hsize=1.0000\hsize\raggedright\arraybackslash}X>{\hsize=1.0000\hsize\raggedleft\arraybackslash}X>{\hsize=1.0000\hsize\raggedleft\arraybackslash}X>{\hsize=1.0000\hsize\raggedleft\arraybackslash}X@{}}
\toprule
 & \multicolumn{2}{c}{\textbf{Numerical comparisons}} & \\
\cmidrule(lr){2-3}
\textbf{Family} & \centering\arraybackslash \textbf{Passing} & \centering\arraybackslash \textbf{Failed} & \centering\arraybackslash \textbf{Runtime errors}\\
\midrule
GDN & 178 / 182 & 4 & 0\\
KDA & 187 / 190 & 3 & 0\\
\bottomrule
\end{tabularx}
\par\vspace{3pt}{\footnotesize\raggedright\textbf{Notes.} Counts refer to the bounded CUDA recurrence and gradient audit. Failed comparisons remain failures under the declared tolerances; zero runtime errors does not establish numerical correctness.\par}
\end{table*}

Acceptance requires finite outputs, elementwise absolute/relative tolerances 0.001/0.02 and relative L2 error at most 0.02, with denominator floor \(10^{-12}\): \(|x_i-o_i| \le 0.001+0.02|o_i|\) with the FP64 oracle as the relative-scale operand, and relative L2 \(\|x-o\|_2 / \max(\|o\|_2,10^{-12})\). KDA gate forward checks use \(2\times10^{-5}\) for all three thresholds. All recurrence outputs pass. Seven zero-decay comparisons fail (Table 11): GDN key/log-decay gradients on both backends, KDA key gradients on both backends and KDA cuDNN state. Their relative L2 errors are below 0.4\%, but individual elements exceed the fixed threshold. Holding the failing fixture inputs fixed and varying log decay over 0, \(-10^{-6}\), −0.001 and −0.01 records 27 failed comparisons out of 112, including failures at negative decays; the largest relative L2 error is 0.5303\%. Eight FP64 directional finite-difference checks of the oracle's gate derivatives pass with maximum absolute discrepancy below 1.44e-10.

\textbf{Checkpoint-derived activations.} Applying the same criterion to prepared recurrence inputs captured from the seed 42 GDN/MoE16 and KDA/MoE16 checkpoints (the first 129 tokens of the first eligible Greek and English validation documents, covering all nine recurrent layers, with fixed-seed random BF16 output derivatives), each family passes 248/252 comparisons, 496/504 overall, with no runtime errors and all outputs passing. GDN fails value gradients at recurrent visit 6 (zero-based) for both documents and both backends; KDA fails fused final state at visit 0 for both documents and key gradients at Greek visit 7 for both backends. Failed tensors contain 1--16 elements beyond the threshold; maximum relative L2 errors are 1.0024\% for GDN and 0.8269\% for KDA. Observed log decay reaches approximately -0.000128 for GDN and -1.70e-8 for KDA.

\textbf{Native next-token loss.} Replacing the random output derivatives with those of native mean next-token cross-entropy over 129 targets, GDN passes 180/180 gradient comparisons and 252/252 overall; KDA passes 178/180 gradient comparisons and 248/252 overall, retaining two fused-state failures. KDA's two gradient failures are the Greek first recurrent layer's value gradients on both backends: 2 of 99,072 elements exceed the elementwise tolerance, with relative L2 error 0.32963\% and maximum absolute error 0.00803563. Maximum gradient relative L2 errors are 1.2075\% for GDN and 0.4924\% for KDA.

These are local operator checks, not end-to-end gradient or optimizer-trajectory comparisons; the gate was not calibrated against a plain sequential BF16 reference, and no effect on optimization or final quality is established.

\section{Complete results}\label{complete-results}

Table 12 reports the held-out test NLL of all 32 selected endpoints.

\begin{table*}[tb]
\caption{Complete fixed held-out evaluation.}\label{tab:layout13}
\footnotesize
\renewcommand{\arraystretch}{1.0}
\begin{tabularx}{\textwidth}{@{}>{\hsize=1.6000\hsize\raggedright\arraybackslash}X>{\hsize=0.3200\hsize\raggedleft\arraybackslash}X>{\hsize=1.0400\hsize\raggedleft\arraybackslash}X>{\hsize=1.0400\hsize\raggedleft\arraybackslash}X@{}}
\toprule
\textbf{Model / numerical control} & \centering\arraybackslash \textbf{Seeds} & \centering\arraybackslash \textbf{Greek test NLL} & \centering\arraybackslash \textbf{English test NLL}\\
\midrule
\rowcolor{groupgray}\multicolumn{4}{@{\hspace{4pt}}l}{\textbf{Primary comparisons (three seeds)}}\\
GQA12 matched mode & 3 & 3.04997 ± 0.00132 & 3.28990 ± 0.00671\\
GQA12 original mode & 3 & 3.05028 ± 0.00180 & 3.28854 ± 0.00403\\
GQA4 & 3 & 3.30434 ± 0.00632 & 3.57620 ± 0.00790\\
GQA4 ×3 shared & 3 & 3.19297 ± 0.00579 & 3.45849 ± 0.00728\\
Hybrid 1:1 & 3 & 3.04014 ± 0.00629 & 3.27513 ± 0.00889\\
Hybrid 3:1 & 3 & 3.04721 ± 0.01422 & 3.28329 ± 0.00487\\
Hybrid MoE16 & 3 & 2.96829 ± 0.00449 & 3.17080 ± 0.00857\\
Hybrid MoE32 & 3 & 2.94604 ± 0.00713 & 3.13998 ± 0.01028\\
KDA/GQA MoE16 & 3 & 2.95105 ± 0.00575 & 3.15397 ± 0.00433\\
\rowcolor{groupgray}\multicolumn{4}{@{\hspace{4pt}}l}{\textbf{Exploratory screens (one seed)}}\\
GDN12 & 1 & 3.16231 & 3.36149\\
GQA NoPE & 1 & 3.15307 & 3.42371\\
GQA half RoPE & 1 & 3.06216 & 3.30735\\
Hybrid conv2 & 1 & 3.05962 & 3.29531\\
Hybrid tail attention & 1 & 3.09204 & 3.32481\\
\bottomrule
\end{tabularx}
\par\vspace{3pt}{\footnotesize\raggedright\textbf{Notes.} Document-segmented held-out test NLL in nats per target (mean ± sample SD); no SD estimated for single-seed screens. All 32 selected endpoints included. Numerical-mode controls are separate groups, not extra independent seeds.\par}
\end{table*}

\begin{figure*}[tb]
\centering
\includegraphics[width=1.0\textwidth]{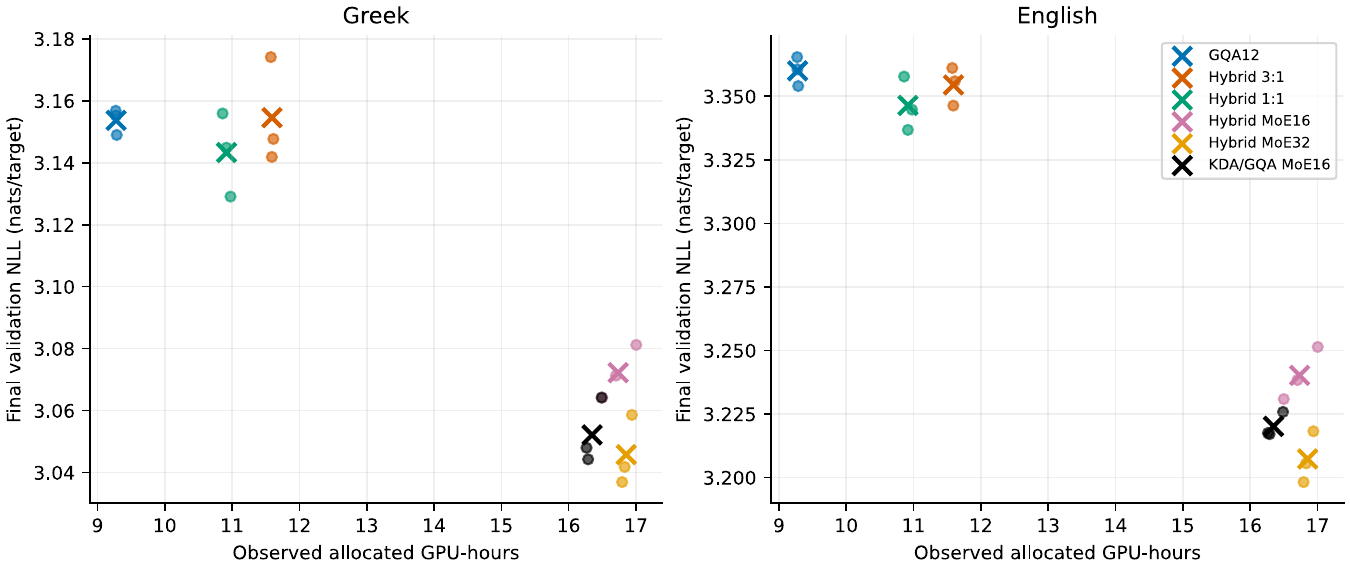}
\caption{Completed three-seed comparisons at equal training targets. Points show 18 individual runs across six three-seed families, including KDA; crosses show group means. Horizontal position is observed allocated training time, not an equal-compute control. Separate language panels share the same runs.}\label{fig:4}
\end{figure*}

\subsection{Perplexity across training seeds}\label{perplexity-across-training-seeds}

Perplexity is \(\exp(\mathrm{NLL})\) for each seed and language on the same held-out targets; Table 13 reports the mean and sample SD of the individual seed perplexities, rather than exponentiating mean NLL.

\begin{table*}[tb]
\caption{Fixed held-out perplexity across training seeds.}\label{tab:layout14}
\footnotesize
\renewcommand{\arraystretch}{1.0}
\begin{tabularx}{\textwidth}{@{}>{\hsize=1.6000\hsize\raggedright\arraybackslash}X>{\hsize=0.3200\hsize\raggedleft\arraybackslash}X>{\hsize=1.0400\hsize\raggedleft\arraybackslash}X>{\hsize=1.0400\hsize\raggedleft\arraybackslash}X@{}}
\toprule
\textbf{Model / numerical control} & \centering\arraybackslash \textbf{Seeds} & \centering\arraybackslash \textbf{Greek test PPL} & \centering\arraybackslash \textbf{English test PPL}\\
\midrule
\rowcolor{groupgray}\multicolumn{4}{@{\hspace{4pt}}l}{\textbf{Primary comparisons (three seeds)}}\\
GQA12 matched mode & 3 & 21.115 ± 0.028 & 26.841 ± 0.180\\
GQA12 original mode & 3 & 21.121 ± 0.038 & 26.804 ± 0.108\\
GQA4 & 3 & 27.231 ± 0.172 & 35.738 ± 0.282\\
GQA4 ×3 shared & 3 & 24.361 ± 0.141 & 31.769 ± 0.232\\
Hybrid 1:1 & 3 & 20.908 ± 0.131 & 26.447 ± 0.235\\
Hybrid 3:1 & 3 & 21.058 ± 0.300 & 26.664 ± 0.130\\
Hybrid MoE16 & 3 & 19.459 ± 0.087 & 23.827 ± 0.204\\
Hybrid MoE32 & 3 & 19.031 ± 0.136 & 23.104 ± 0.238\\
KDA/GQA MoE16 & 3 & 19.126 ± 0.110 & 23.429 ± 0.101\\
\rowcolor{groupgray}\multicolumn{4}{@{\hspace{4pt}}l}{\textbf{Exploratory screens (one seed)}}\\
GDN12 & 1 & 23.625 & 28.832\\
GQA NoPE & 1 & 23.408 & 30.683\\
GQA half RoPE & 1 & 21.374 & 27.313\\
Hybrid conv2 & 1 & 21.319 & 26.986\\
Hybrid tail attention & 1 & 22.022 & 27.794\\
\bottomrule
\end{tabularx}
\par\vspace{3pt}{\footnotesize\raggedright\textbf{Notes.} Perplexity is exp(NLL) for each seed. Entries show mean ± sample SD of seed perplexities, not exp(mean NLL). Single-seed screens have no SD. Same fixed test pools as Table 12; lower is better.\par}
\end{table*}

\subsection{Paired contrasts with dispersion and document-bootstrap intervals}\label{paired-contrasts-with-dispersion-and-document-bootstrap-intervals}

Table 14 lists every paired contrast; the bootstrap is described in Section 3.4.

\begin{table*}[tb]
\caption{Paired contrasts with dispersion and document-bootstrap intervals.}\label{tab:layout15}
\footnotesize
\renewcommand{\arraystretch}{1.0}
\begin{tabularx}{\textwidth}{@{}>{\hsize=0.9500\hsize\raggedright\arraybackslash}X>{\hsize=0.7500\hsize\raggedleft\arraybackslash}X>{\hsize=0.7500\hsize\raggedleft\arraybackslash}X>{\hsize=1.2750\hsize\raggedleft\arraybackslash}X>{\hsize=1.2750\hsize\raggedleft\arraybackslash}X@{}}
\toprule
\textbf{Contrast} & \centering\arraybackslash \textbf{Greek validation} & \centering\arraybackslash \textbf{English validation} & \centering\arraybackslash \textbf{Greek test} & \centering\arraybackslash \textbf{English test}\\
\midrule
\rowcolor{groupgray}\multicolumn{5}{@{\hspace{4pt}}l}{\textbf{Expert capacity}}\\
GDN/MoE16 − GQA12 & −0.0814 ± 0.0064 (3/3) & −0.1198 ± 0.0094 (3/3) & −0.0820 {[}−0.0857, −0.0784{]} (3/3) & −0.1177 {[}−0.1213, −0.1141{]} (3/3)\\
GDN/MoE32 − GDN/MoE16 & −0.0265 ± 0.0035 (3/3) & −0.0329 ± 0.0003 (3/3) & −0.0223 {[}−0.0236, −0.0208{]} (3/3) & −0.0308 {[}−0.0321, −0.0295{]} (3/3)\\
GDN/MoE16 − hybrid 3:1 & −0.0823 ± 0.0092 (3/3) & −0.1143 ± 0.0040 (3/3) & −0.0789 {[}−0.0820, −0.0758{]} (3/3) & −0.1125 {[}−0.1155, −0.1093{]} (3/3)\\
\rowcolor{groupgray}\multicolumn{5}{@{\hspace{4pt}}l}{\textbf{Dense mixing}}\\
Hybrid 3:1 − GQA12 & +0.0009 ± 0.0156 (2/3) & −0.0056 ± 0.0053 (2/3) & −0.0031 {[}−0.0047, −0.0015{]} (2/3) & −0.0052 {[}−0.0065, −0.0041{]} (3/3)\\
Hybrid 1:1 − hybrid 3:1 & −0.0113 ± 0.0123 (2/3) & −0.0080 ± 0.0041 (3/3) & −0.0071 {[}−0.0086, −0.0057{]} (2/3) & −0.0082 {[}−0.0093, −0.0070{]} (3/3)\\
Hybrid 1:1 − GQA12 & −0.0104 ± 0.0152 (2/3) & −0.0136 ± 0.0095 (3/3) & −0.0101 {[}−0.0113, −0.0090{]} (3/3) & −0.0134 {[}−0.0147, −0.0121{]} (3/3)\\
\rowcolor{groupgray}\multicolumn{5}{@{\hspace{4pt}}l}{\textbf{Shared depth}}\\
Shared GQA4 ×3 − GQA4 & −0.1087 ± 0.0097 (3/3) & −0.1159 ± 0.0128 (3/3) & −0.1114 {[}−0.1144, −0.1083{]} (3/3) & −0.1177 {[}−0.1197, −0.1159{]} (3/3)\\
Shared GQA4 ×3 − GQA12 matched & +0.1482 ± 0.0022 (0/3) & +0.1725 ± 0.0124 (0/3) & +0.1430 {[}+0.1398, +0.1462{]} (0/3) & +0.1686 {[}+0.1649, +0.1721{]} (0/3)\\
\rowcolor{groupgray}\multicolumn{5}{@{\hspace{4pt}}l}{\textbf{Recurrent family}}\\
KDA/MoE16 − GDN/MoE16 & −0.0201 ± 0.0031 (3/3) & −0.0201 ± 0.0059 (3/3) & −0.0172 {[}−0.0187, −0.0157{]} (3/3) & −0.0168 {[}−0.0179, −0.0157{]} (3/3)\\
\bottomrule
\end{tabularx}
\par\vspace{3pt}{\footnotesize\raggedright\textbf{Notes.} Candidate minus baseline NLL, nats per target; negative favors the candidate. Validation: mean ± sample SD of paired-seed differences. Test: mean with 95\% paired document-bootstrap interval (2,000 resamples), which excludes seed variation. Parentheses: improved paired seeds.\par}
\end{table*}

\subsection{Greek held-out contrasts by source component}\label{greek-held-out-contrasts-by-source-component}

The Greek test components contain 285 (synthetic mathematics), 733 (stories), 555 (web) and 62 (reference and other) documents. Table 15 splits each Greek test contrast by component.

\begin{table*}[tb]
\caption{Greek held-out contrasts by source component.}\label{tab:layout16}
\footnotesize
\renewcommand{\arraystretch}{1.0}
\begin{tabularx}{\textwidth}{@{}>{\hsize=1.4000\hsize\raggedright\arraybackslash}X>{\hsize=0.9000\hsize\raggedleft\arraybackslash}X>{\hsize=0.9000\hsize\raggedleft\arraybackslash}X>{\hsize=0.9000\hsize\raggedleft\arraybackslash}X>{\hsize=0.9000\hsize\raggedleft\arraybackslash}X@{}}
\toprule
\textbf{Greek test contrast} & \centering\arraybackslash \textbf{Synthetic mathematics} & \centering\arraybackslash \textbf{Stories} & \centering\arraybackslash \textbf{Web} & \centering\arraybackslash \textbf{Reference and other}\\
\midrule
\rowcolor{groupgray}\multicolumn{5}{@{\hspace{4pt}}l}{\textbf{Expert capacity}}\\
GDN/MoE16 − GQA12 & −0.0314 (3/3) & −0.1090 (3/3) & −0.1073 (3/3) & −0.1036 (3/3)\\
GDN/MoE32 − GDN/MoE16 & −0.0088 (3/3) & −0.0316 (3/3) & −0.0282 (3/3) & −0.0267 (3/3)\\
GDN/MoE16 − hybrid 3:1 & −0.0371 (3/3) & −0.1069 (3/3) & −0.0977 (3/3) & −0.0959 (3/3)\\
\rowcolor{groupgray}\multicolumn{5}{@{\hspace{4pt}}l}{\textbf{Dense mixing}}\\
Hybrid 3:1 − GQA12 & +0.0057 (2/3) & −0.0021 (1/3) & −0.0096 (3/3) & −0.0078 (2/3)\\
Hybrid 1:1 − hybrid 3:1 & −0.0137 (2/3) & −0.0079 (2/3) & −0.0024 (2/3) & +0.0005 (2/3)\\
Hybrid 1:1 − GQA12 & −0.0079 (1/3) & −0.0101 (3/3) & −0.0120 (3/3) & −0.0073 (2/3)\\
\rowcolor{groupgray}\multicolumn{5}{@{\hspace{4pt}}l}{\textbf{Shared depth}}\\
Shared GQA4 ×3 − GQA4 & −0.1184 (3/3) & −0.1320 (3/3) & −0.0983 (3/3) & −0.1082 (3/3)\\
Shared GQA4 ×3 − GQA12 matched & +0.1164 (0/3) & +0.1729 (0/3) & +0.1497 (0/3) & +0.1618 (0/3)\\
\rowcolor{groupgray}\multicolumn{5}{@{\hspace{4pt}}l}{\textbf{Recurrent family}}\\
KDA/MoE16 − GDN/MoE16 & −0.0273 (3/3) & −0.0141 (3/3) & −0.0107 (3/3) & −0.0221 (3/3)\\
\bottomrule
\end{tabularx}
\par\vspace{3pt}{\footnotesize\raggedright\textbf{Notes.} Mean paired test difference within each Greek source component, nats per target; parentheses give improved paired seeds. Secondary descriptive analysis added after scoring.\par}
\end{table*}

\subsection{Supplementary output-gate comparison}\label{supplementary-output-gate-comparison}

Two matched sigmoid-output KDA runs (seeds 42 and 43) passed endpoint integrity, data-lineage and restoration checks. At seed 42, sigmoid versus SiLU Greek/English validation NLL is 3.03718/3.20808 versus 3.06419/3.22581; at seed 43 it is 3.01883/3.18749 versus 3.04426/3.21707. Mean paired differences are −0.02622 Greek and −0.02365 English nats/target, with sample SDs 0.00112 and 0.00838. This two-seed screen was not scored on the held-out test and does not reproduce Kimi Linear's low-rank gate or MLA backbone; it is a direction for follow-up.

\subsection{Gradient clipping by group}\label{gradient-clipping-by-group}

Table 16 summarizes post-warmup gradient norms; ranges span the seeds of each group.

\begin{table*}[tb]
\caption{Gradient clipping by group.}\label{tab:layout17}
\footnotesize
\renewcommand{\arraystretch}{1.0}
\begin{tabularx}{\textwidth}{@{}>{\hsize=1.6200\hsize\raggedright\arraybackslash}X>{\hsize=0.4800\hsize\raggedleft\arraybackslash}X>{\hsize=1.0200\hsize\raggedleft\arraybackslash}X>{\hsize=0.8400\hsize\raggedleft\arraybackslash}X>{\hsize=1.0200\hsize\raggedleft\arraybackslash}X>{\hsize=1.0200\hsize\raggedleft\arraybackslash}X@{}}
\toprule
\textbf{Group} & \centering\arraybackslash \textbf{Runs} & \centering\arraybackslash \textbf{Clipped updates, \%} & \centering\arraybackslash \textbf{Median norm} & \centering\arraybackslash \textbf{99th-percentile norm} & \centering\arraybackslash \textbf{Maximum norm}\\
\midrule
\rowcolor{groupgray}\multicolumn{6}{@{\hspace{4pt}}l}{\textbf{Primary groups (three seeds)}}\\
GQA12, original mode & 3 & 99.10--99.28 & 3.02--3.20 & 5.35--5.41 & 109.43\\
GQA12, matched mode & 3 & 99.10--99.24 & 2.99--3.11 & 5.03--5.05 & 82.41\\
Hybrid 3:1 & 3 & 99.89--99.92 & 2.78--2.89 & 4.31--4.80 & 49.34\\
Hybrid 1:1 & 3 & 99.85--99.91 & 2.78--2.92 & 4.36--4.73 & 41.02\\
GDN/MoE16 & 3 & 99.10--99.22 & 2.61--2.81 & 4.06--4.65 & 35.26\\
GDN/MoE32 & 3 & 98.71--98.91 & 2.67--2.73 & 4.07--4.11 & 83.75\\
KDA/MoE16 & 3 & 98.94--98.98 & 2.68--2.76 & 3.98--4.22 & 183.40\\
GQA4 & 3 & 99.98--99.99 & 3.36--3.57 & 6.78--8.94 & 87.07\\
GQA4 ×3 shared & 3 & 99.98 & 3.23--3.39 & 6.61--6.89 & 171.25\\
\rowcolor{groupgray}\multicolumn{6}{@{\hspace{4pt}}l}{\textbf{Exploratory screens (one seed)}}\\
GDN12 & 1 & 99.96 & 2.99 & 5.06 & 50.10\\
GQA NoPE & 1 & 99.53 & 3.11 & 5.60 & 384.58\\
GQA half RoPE & 1 & 99.12 & 3.17 & 5.43 & 145.61\\
Hybrid conv2 & 1 & 99.89 & 3.20 & 5.75 & 162.84\\
Hybrid tail attention & 1 & 99.80 & 2.91 & 4.55 & 31.59\\
\bottomrule
\end{tabularx}
\par\vspace{3pt}{\footnotesize\raggedright\textbf{Notes.} Post-warmup logged gradient norms before clipping at 1. Ranges span seeds within a group.\par}
\end{table*}

\begin{figure*}[p]
\centering
\includegraphics[width=1.0\textwidth]{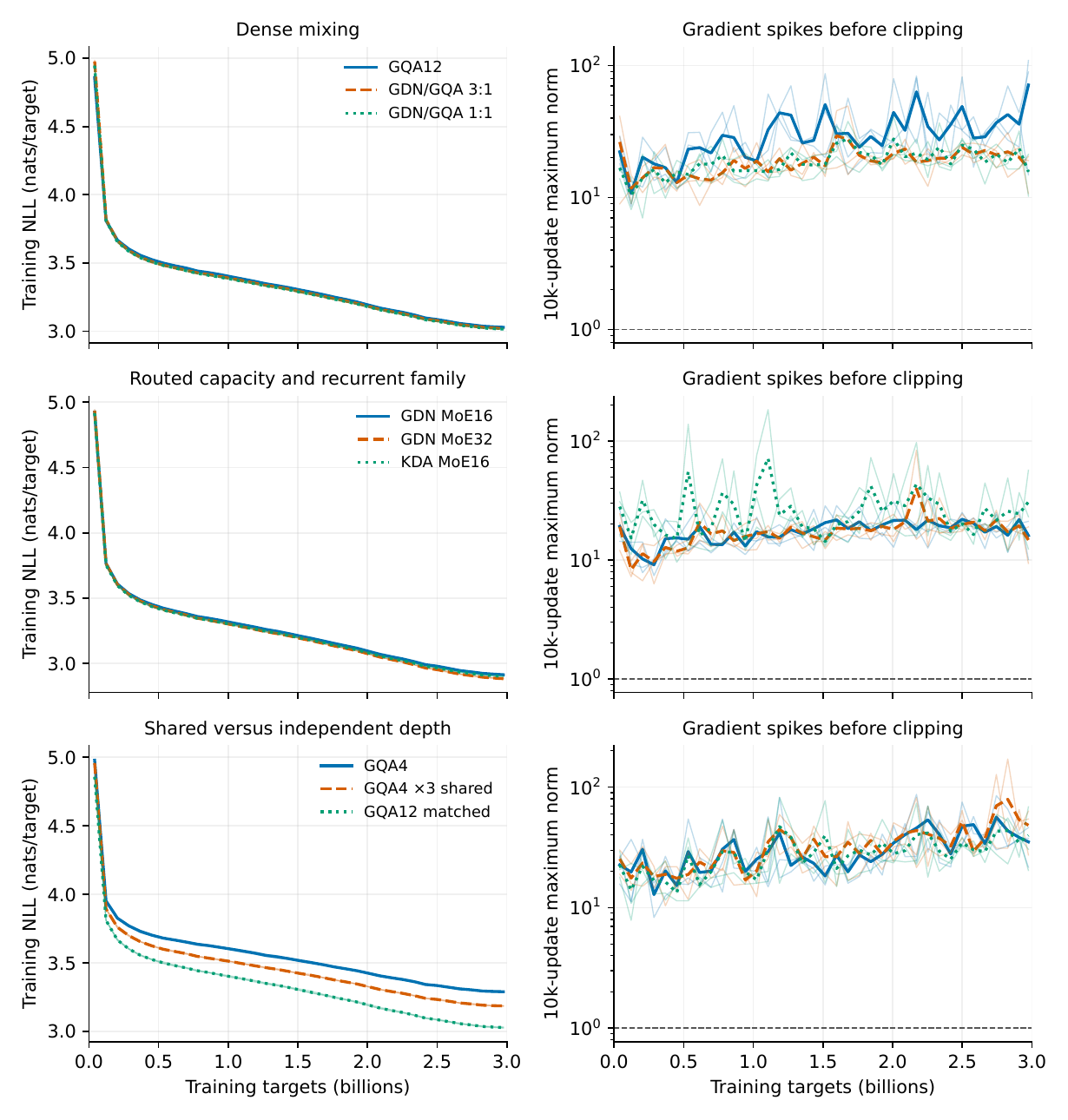}
\caption{Training loss and pre-clipping gradient norms for 27 runs across nine primary groups. Thin lines show individual seeds; thick lines average three seed summaries. Loss panels use 10,000-update means; gradient panels use within-bin maxima on a logarithmic scale. The horizontal reference is the clipping threshold of 1. Appendix E.6 defines aggregation and the separate excursion diagnostic. Finite completion under clipping does not establish unclipped stability.}\label{fig:5}
\end{figure*}

\subsection{Training trajectories and gradient excursions}\label{training-trajectories-and-gradient-excursions}

The audit covers the complete logs of all 32 selected runs, not interrupted attempts at a separate 12-billion-target run or qualification pilots. The logs contain no out-of-memory, traceback or non-finite-stop markers. The largest logged norm is 384.5843 (exploratory NoPE run). The small batch of 8,192 targets per update may contribute to the high norms.

As a descriptive spike diagnostic (Figure 5 shows the trajectories), a post-warmup update is flagged when its logged norm exceeds five times the median of the preceding 100 post-warmup norms. This retrospective threshold flags 4,089 of 11,598,336 eligible updates across all 32 runs. Spikes remain finite; spike counts cannot rank intrinsic stability.
\end{document}